\documentclass[runningheads]{llncs}
\usepackage{graphicx}

\usepackage{tikz}
\usepackage{comment}
\usepackage{amsmath,amssymb} 
\usepackage{color}
\usepackage[pagebackref,breaklinks,colorlinks]{hyperref}
\usepackage{subcaption}
\usepackage{booktabs}
\usepackage{cite}
\usepackage{comment}

\usepackage[accsupp]{axessibility}  

\begin{document}
\pagestyle{headings}
\mainmatter

\title{SSBNet: Improving Visual Recognition Efficiency by Adaptive Sampling} 
\titlerunning{SSBNet: Improving Visual Recognition Efficiency by Adaptive Sampling}

\author{Ho Man Kwan\index{Kwan, Ho Man}, Shenghui Song}
\author{
	Ho Man Kwan\index{Kwan, Ho Man} \and
	Shenghui Song\index{Song, Shenghui}
}
\authorrunning{H.M. Kwan \and S.H. Song}
\institute{
	The Hong Kong University of Science and Technology \\
	\email{hmkwan@connect.ust.hk} \quad \email{eeshsong@ust.hk}
}
\maketitle

\begin{abstract}
Downsampling is widely adopted to achieve a good trade-off between accuracy and latency for visual recognition. Unfortunately, the commonly used pooling layers are not learned, and thus cannot preserve important information. As another dimension reduction method, adaptive sampling weights and processes regions that are relevant to the task, and is thus able to better preserve useful information. However, the use of adaptive sampling has been limited to certain layers. 
In this paper, we show that using adaptive sampling in the building blocks of a deep neural network can improve its efficiency.
In particular, we propose SSBNet which is built by inserting sampling layers repeatedly into existing networks like ResNet. Experiment results show that the proposed SSBNet can achieve competitive image classification and object detection performance on ImageNet and COCO datasets. For example, the SSB-ResNet-RS-200 achieved 82.6\% accuracy on ImageNet dataset, which is 0.6\% higher than the baseline ResNet-RS-152 with a similar complexity. Visualization shows the advantage of SSBNet in allowing different layers to focus on different positions, and ablation studies further validate the advantage of adaptive sampling over uniform methods.
\end{abstract}

\section{Introduction}
\label{sec:intro}

\begin{figure}[t]
	\begin{subfigure}[t]{0.475\textwidth}
		\centering
		\includegraphics[width=5cm,trim={0 0 0 1.2cm 0},clip]{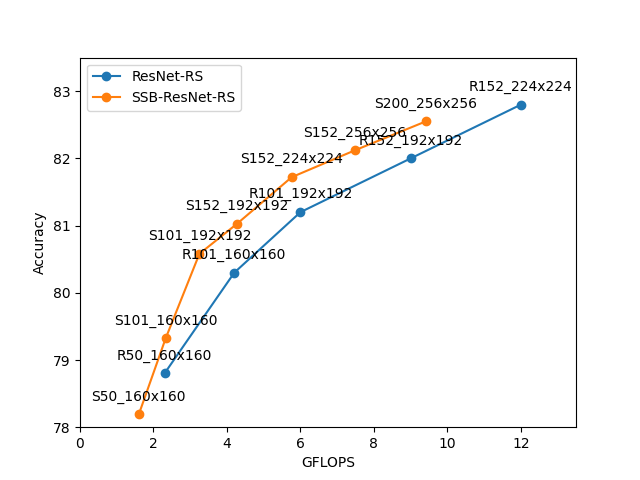}
		\caption{}
		\label{fig:resnetssbresnet}
	\end{subfigure}
	\begin{subfigure}[t]{0.475\textwidth}
		\centering
		\includegraphics[width=5cm,trim={0 0 0 1.2cm 0},clip]{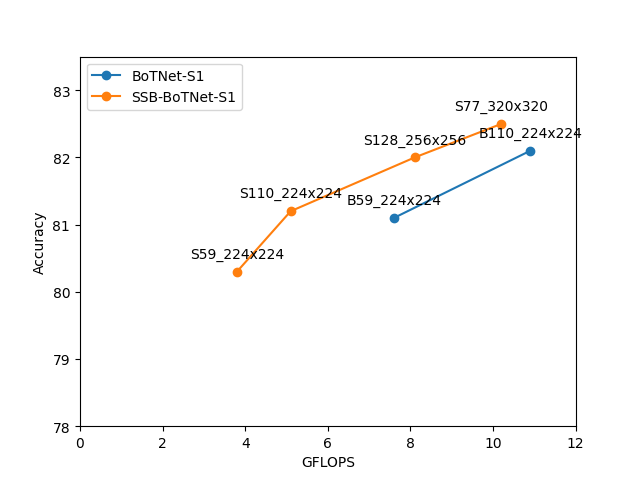}
		\caption{}
		\label{fig:botnetssbbotnet}
	\end{subfigure}
	\caption{Comparison between the SSB-ResNet-RS and ResNet-RS\cite{bello2021revisiting}, SSB-BoTNet-S1 and BoTNet-S1\cite{srinivas2021bottleneck} on ImageNet \cite{ILSVRC15} dataset. The proposed SSB-ResNet-RS/SSB-BoTNet-S1 outperforms ResNet-RS and BoTNet-S1 in terms of accuracy to FLOPS ratio. Results of ResNet-RS are from the original paper, where the results of BoTNet-S1 are from reimplementation. See section \ref{subsec:image classification}.}
	\label{fig:resnet_botnet_ssb}
\end{figure}

Deep learning models such as convolutional neural networks (CNNs)\cite{krizhevsky2012imagenet, szegedy2015going,  simonyan15, he2016deep} and Transformers\cite{dosovitskiy2021an} have made unprecedented successes in computer vision. However, achieving efficient inference with stringent latency constraints in real-world applications is very challenging. To obtain a good trade-off between accuracy and latency, downsampling is normally used to reduce the number of operations. Most existing CNNs\cite{krizhevsky2012imagenet, szegedy2015going,  simonyan15, he2016deep} perform downsampling between stages, coupled with the increase in channel dimension to balance the representation power and computational cost. Typical downsampling operations include strided average/max pooling and convolutions \cite{krizhevsky2012imagenet, szegedy2015going,  simonyan15, he2016deep, peng2018accelerating,li2019hbonet}, which are uniformly applied in the spatial dimension.

Besides uniform sampling, there are also non-uniform or adaptive approaches \cite{jaderberg2015spatial, recasens2018learning, zheng2019looking, jin2022learning, thavamani2021fovea, dai2017deformable}, with which different transformations including zooming, shifting, and deforming can be utilized to selectively focus on the important regions during downsampling. However, the use of adaptive sampling has been limited to certain layers and its application in backbone networks has not been well investigated. Backbone networks are usually pre-trained in some large scale datasets, which are agnostic to the end task like object detection \cite{zhao2019object}. The challenge for applying adaptive sampling in backbone networks lies in the possible information loss. In particular, later layers cannot access pixels that were skipped by earlier layers, which is quite possible due to the stacking of sampling layers.  

Another approach to improve efficiency is to reduce the number of channels. In ResNet\cite{he2016deep}, the bottleneck layers reduce the channel dimension by using a $1 \times 1$ convolution and then perform a costly $3 \times 3$ convolution on the low dimensional features to reduce computational complexity. After that, another $1 \times 1$ convolution is utilized to restore the dimension and match the shortcut connection. It is noteworthy that residual networks can preserve informative features after the bottleneck operations because the shortcut connection allows signal to bypass the bottleneck.

The bottleneck structure with shortcut connection can be utilized to enable adaptive sampling on backbone networks. In this paper, we propose Saliency Sampling Bottleneck Network (SSBNet), which applies saliency sampler \cite{recasens2018learning,zheng2019looking} in a bottleneck structure to reduce the spatial dimension before costly operations. An inverse operation is then used to restore the feature maps to match the spatial structure of the input that passes through the shortcut connection. Like other bottleneck structures, computationally expensive operations like convolution are applied in a very compact space to save computations. There are two major advantages for applying adaptive sampling over the bottleneck structure. First, by zooming into important regions, SSBNet can better extract features than uniform downsampling. More importantly, with the shortcut connection, each intermediate layer with adaptive sampling can focus on different regions of the feature maps in a very deep network, without loss of information.

In the experiments, we built SSBNets by inserting lightweight convolutional layers and samplers into existing networks to estimate the saliency map and perform down/upsampling. The results in Figure \ref{fig:resnet_botnet_ssb} show that SSBNet can achieve better accuracy/FLOPS ratio than ResNet-RS\cite{bello2021revisiting} and BoTNet-S1\cite{srinivas2021bottleneck}. For example, 
with only 4.4\% more FLOPS, the SSB-ResNet-RS-200 with input size of $256\times256$ achieved 82.6\% accuracy on ImageNet dataset\cite{ILSVRC15}, which is 0.6\% higher than the baseline ResNet-RS-152 with input size of $192\times192$. The SSB-BoTNet-S1-77 with input size of $320\times320$ obtained 82.5\% accuracy, which is 0.4\% higher than BoTNet-S1-110 with input size of $224\times224$, and required 6\% less computation.
The contributions of this paper include:
\begin{itemize}

	\item We investigate the use of adaptive sampling in the building blocks of a deep neural network. By applying adaptive sampling in the bottleneck structure, we propose SSBNet which can be utilized as a backbone network and trained in an end-to-end manner. Note that existing networks only utilized adaptive sampling in specific tasks, where a pre-trained backbone is required for feature extraction. 
	\item We show that the proposed SSBNet can achieve better image classification and object detection performance than the baseline models, and visualize its capability in adaptively sampling different locations at different layers. 
	\item Experiment results and ablation studies validate the advantage of adaptive sampling over uniform sampling. The result in this paper may lead to a new direction of research on network architecture.
\end{itemize}

\section{Related Works}
\label{sec:related works}
In the following, we explain the connection between the proposed SSBNet and existing works, and highlight the innovation.

\subsubsection{Attention Mechanisms.}  Different types of attention mechanisms have been explored in computer vision tasks. One category of work utilizes attention mechanism to predict a softmask that scales the feature maps. Squeeze-and-Excitation\cite{hu2018squeeze} utilizes global context to refine the channel dimension. CBAM\cite{woo2018cbam} uses attention mask to emphasize the important spatial positions.

Besides improving feature maps, another direction of research applies attention as a stand-alone layer that can extract features and act as a replacement for the convolutional layer. Stand-alone self-attention \cite{ramachandran2019stand} replaces the spatial convolutional layer to efficiently increase the receptive field. Vision Transformer \cite{dosovitskiy2021an} adapts the Transformer\cite{vaswani2017attention} structure and takes non-overlapped patches as individual tokens, instead of a map representation that is normally used in vision tasks.

The proposed SSBNet follows the first approach and inserts attention layers to improve efficiency. However, instead of utilizing attention to scale features, SSBNet performs weighted downsampling by the attention map to save computations.

\subsubsection{Adaptive Sampling.} There are some works\cite{jaderberg2015spatial, recasens2018learning, zheng2019looking, jin2022learning, thavamani2021fovea, dai2017deformable} that perform adaptive geometric sampling on the images or feature maps rather than scaling the features, as is done by attention mechanisms. Spatial transformer network\cite{jaderberg2015spatial} uses localization network to predict transformation parameters and performs geometric transformation on the image or feature maps. Saliency sampler\cite{recasens2018learning} applies saliency map estimator to compute the attention map and distorts the input based on this map. Trilinear attention sampling network (TASN)\cite{zheng2019looking} applies trilinear attention to compute the attention map and uses the map to perform sampling in a less distorted way. The sampling mechanism of the proposed SSBNet is inspired by TASN, but with two major differences: 1) SSBNet can be used as a backbone and trained end-to-end, but TASN requires a pre-trained backbone; 2) SSBNet performs different sampling at different layers to extract useful features, where TASN only performs sampling once on the image input.

There are only very few works that apply adaptive sampling in the backbone network, which is the core feature extractor for computer vision tasks. One of the exceptions is the Deformable convolutional neural network (DCNN) \cite{dai2017deformable}. DCNN computes the sampling offset to deform the sampling grid of convolutions and RoI poolings, which provides significant performance improvement for object detection and semantic segmentation tasks. Different from DCNN which deforms the convolutions and RoI poolings, SSBNet samples the feature map into a lower dimension to improve efficiency.

In summary, the proposed SSBNet utilizes adaptive sampling in most of its building blocks and allows different sampling at different layers, where most existing works only perform adaptive sampling several times. SSBNet can be used as a backbone network for different tasks like classification and object detection. 

\subsubsection{Dimension Reduction.} Dimension reduction is commonly used in different architectures. Reducing spatial dimensions can save a large amount of computation and increase the effective receptive field of the convolution operations. For example, many CNNs reduce the spatial dimension when they increase the number of channels\cite{krizhevsky2012imagenet,simonyan15,szegedy2015going,he2016deep}. 

There are networks that temporarily reduce the channel dimension. Inception\cite{szegedy2015going} applies $1\times1$ convolutions to reduce the channel dimension and lower the cost of the following $3\times3$ and $5\times5$ convolutions. ResNet\cite{he2016deep} has a bottleneck layer design, which reduces the number of channels before the $3\times3$ convolution and restores the channel dimension afterwards.

There are also applications of the bottleneck structure in the spatial dimension. Spatial bottleneck\cite{peng2018accelerating} replaces spatial convolution by a pair of strided convolution and deconvolution to reduce the sampling rate and achieve speed-up. HBONet\cite{li2019hbonet} utilizes depthwise convolution and bilinear sampling to perform down/upsampling, where the costly operations are applied in between. 

The proposed SSBNet has a similar structure as HBONet, but utilizes adaptive sampling instead of strided convolution or pooling for downsampling. Furthermore, the adaptive sampling can perform spatial transformation like zooming, which could better preserve useful information for feature extraction.

\section{Methodology}
\label{sec:methodology}

\begin{figure}[t]
	\centering
	\includegraphics[trim={0 0.1cm 0 0.1cm},clip,height=6cm]{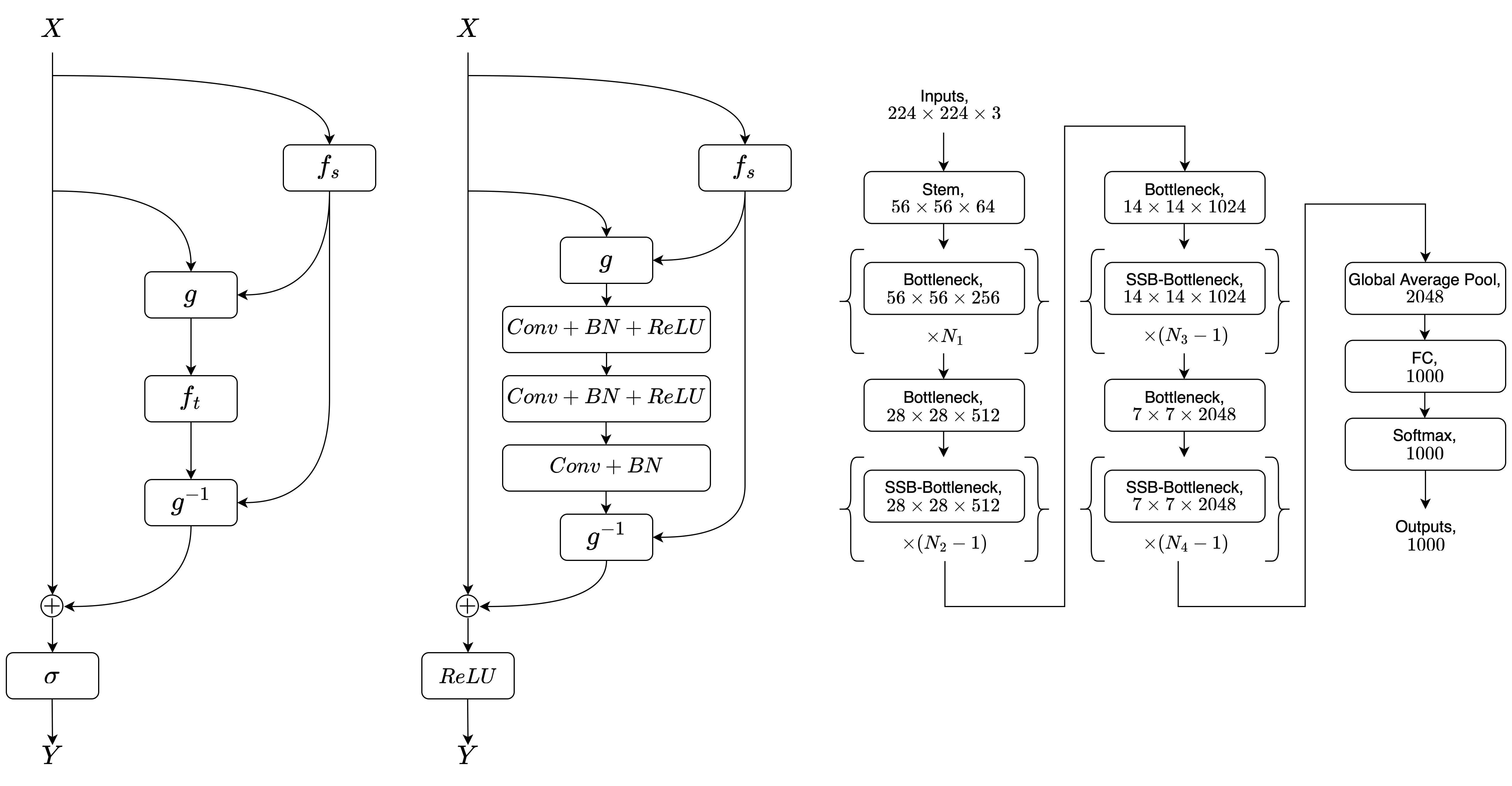}
	\caption{Left: The structure of SSB layer. Middle: The instantiation of SSB layer built from the bottleneck layer of ResNet\cite{he2016deep}. Right: SSB-ResNet, where $N_1, N_2, N_3, N_4$ follow the configurations of ResNet\cite{he2016deep}. } 
	\label{fig:ssnssbb}
\end{figure}

In this section, we first introduce the SSBNet and then present the details of the building block for SSBNet, i.e. the SSB layer.

\subsection{Saliency Sampling Bottleneck Networks}
\label{subsec:ssbnet}

Since the focus of this work is to apply adaptive sampling to improve network efficiency, we modify existing networks to reduce the searching space. To build SSBNet, we insert samplers to the building blocks of the original model such as ResNet\cite{he2016deep}. To this end, we only need to determine the sampling size and the position to insert the sampler. In the experiments, we follow the standard approach that shares configuration in a group of building blocks, i.e. same sampling size in one group. Without native implementation of the sampler, some earliest groups that have high spatial dimension will significantly slow down the training. So we skip those earliest groups.

We also skip the first block of each group, i.e. the block that reduces the spatial dimension and increases the number of channels, due to the fact that they usually utilize shortcut connection with strided pooling or/and $1\times1$ convolution for downsampling\cite{he2016deep, he2019bag, bello2021revisiting}, or do not contain shortcut connection\cite{tan2019efficientnet}. Thus, adding samplers to the first block of each group could lead to loss of information. For example, SSB-ResNet is shown in Figure \ref{fig:ssnssbb} (right).

\subsection{Saliency Sampling Bottleneck Layer}
\label{subsec:ssb layer}

The SSB layer is the main building block of the SSBNet and is constructed by wrapping a set of layers with the samplers. The SSB layer has a similar structure as the bottleneck layer from ResNet\cite{he2016deep}, with two branches, i.e., the shortcut branch and the residual branch. The function of the two branches are similar to those in ResNet. Specifically, the shortcut branch passes the signal to the higher level layer and the residual branch performs operations to extract features. The difference is that the residual branch in the SSB layer adaptively samples features in the spatial dimension, but the bottleneck layer of ResNet reduces the channel dimension, and both of them perform the most costly operations in the reduced space. Figure \ref{fig:ssnssbb} (left) shows the structure of the SSB layer. Next, we introduce the key operations of the SSB layer.

\textbf{Saliency Map}: Given the input feature map $X\in \mathbb{R}^{H_{in} \times W_{in} \times D}$, where $H_{in}$, $W_{in}$, $D$ denote the height, the weight and the number of channels, respectively, we first compute the saliency map by $S = f_s(X)$,
where $S$ has dimensions of $H_{in} \times W_{in}$. There are many possible choices of $f_s$. In this paper, we use a $1 \times 1$ convolutional layer with one filter, a batch normalization layer\cite{ioffe2015batch} and the sigmoid activation, followed by a reshape operation which change the map of  $H_{in} \times W_{in} \times 1$ to a 2D matrix of $H_{in} \times W_{in}$. The whole process has negligible overhead in the number of operations and parameters. To stabilize the training, we always initialize the scaling weight $\gamma$ in the batch normalization layer to zero, such that the network performs uniform sampling at the beginning.

\textbf{Sampling Output Computation}: For a target sampling size of ${H_r \times W_r}$, we compute the sampling output $X^r = g(X, S)$ with $X^r \in \mathbb{R}^{H_r \times W_r \times D}$.
Our approach is close to TASN\cite{zheng2019looking}. Specifically, we apply inverse transform to convert the saliency map into the weights of sampling, where features having higher scores in the saliency map will be sampled with a larger weight into the output feature maps. Unlike TASN, our implementation does not involve bilinear sampling\cite{jaderberg2015spatial}. Instead, we directly compute the sampling weights between the input and output pixels.

To compute $X^r$ with a saliency map $S \in \mathbb{R} ^{H_{in} \times W_{in}}$, we first obtain the elements of the saliency vectors
$S^y \in \mathbb{R} ^{H_{in}}$ 
and
$S^x \in \mathbb{R} ^{W_{in}}$
as
\begin{equation}
	\begin{gathered}
		S^y_j = \frac{\sum_{w=1}^{W_{in}}{S_{j, w}}}{\sum_{h=1}^{H_{in}}\sum_{w=1}^{W_{in}}{S_{h, w}}}
		\quad \forall 1 \leq j \leq H_{in}
	\end{gathered}
	\label{eq:saliency vector S^H}
\end{equation}
and
\begin{equation}
	\begin{gathered}
		S^x_i = \frac{\sum_{h=1}^{H_{in}}{S_{h, i}}}{\sum_{h=1}^{H_{in}}\sum_{w=1}^{W_{in}}{S_{h, w}}}
		\quad \forall 1 \leq i \leq W_{in}.
	\end{gathered}
	\label{eq:saliency vector S^W}
\end{equation}
Note that both $S^y$ and $S^x$ are normalized.

We also compute uniform vectors, $U^y \in \mathbb{R} ^{H_{r}}$ and $U^x \in \mathbb{R} ^{W_{r}}$, where
\begin{equation}	
	U^y_j = \frac{1}{H_r} \quad \forall  1 \leq j \leq H_r
	\label{eq:uniform vector U^y}
\end{equation}
\begin{equation}
	U^x_i = \frac{1}{W_r} \quad \forall  1 \leq i \leq W_r.
	\label{eq:uniform vector U^x}
\end{equation}
Then, we calculate the cumulative sums $C^{S^y}$, $C^{S^x}$, $C^{U^y}$ and $C^{U^x}$. For example, in the y-axis, we first compute
\begin{equation}
	C^{S^y}_{j} = \sum_{h=1}^{j-1}{S^y_{h}}  \quad \forall  1 \leq j \leq H_{in} + 1
	\label{eq:cumulative sum S^x}
\end{equation}
\begin{equation}
	C^{U^y}_{j} = \sum_{h=1}^{j-1}{U^y_{h}}  \quad \forall  1 \leq j \leq H_r + 1
	\label{eq:cumulative sum U^x}
\end{equation}
and then the sampling weights can be determined as
\begin{equation}
	\begin{gathered}
		G^y_{i, j} = max(min(C^{S^y}_{j+1}, C^{U^y}_{i+1}) - max(C^{S^y}_{j}, C^{U^y}_{i}), 0)
		\\\forall  1 \leq j \leq H_{in}, 1 \leq i \leq H_r.
	\end{gathered}
	\label{eq:saliency sampler weght G^y}
\end{equation}
The weight matrix in the x-axis, $G^x$, can be computed similarly. Weight matrices $G^y$ and $G^x$ have dimensions of $H_r \times H_{in}$ and $W_r \times W_{in}$, respectively.

Finally, we can compute the sampling output $X^r$ by
\begin{equation}
	\begin{gathered}
		X^r_{i, j, d} = \sum_{h=1}^{H_{in}}    \sum_{w=1}^{W_{in}}    H_r  W_r G^y_{i, h}    G^x_{j, w} X_{h, w, d}
		\\\forall  1 \leq i \leq H_r, 1 \leq j \leq W_r, 1 \leq d \leq D.
	\end{gathered}
	\label{eq:saliency sampler sampling}
\end{equation}
We applied scaling with a factor of $H_r  W_r$, such that the average value of the output map is independent of the sampling size.

\textbf{Feature Extraction}: After computing the sampled feature maps $X^r$, we can extract features by costly operations like convolutions with $Y^r = f_t(X^r)$.
When building SSBNet from existing networks with shortcut connections\cite{he2016deep}, we use the original residual branch as $f_t$. Figure \ref{fig:ssnssbb} (middle) shows the SSB layer built from the (channel) bottleneck layer of ResNet.

\textbf{Inverse Sampling}: After the feature extraction stage, an inverse sampling is applied to restore the spatial dimension. For that purpose, we apply the same sampling method, except that the transposed weight matrices, i.e. $(G^y)^T$ and $(G^x)^T$, are utilized. Together with the shortcut connection and the activation function $\sigma$, the final output of the SSB layer can be expressed as
\begin{equation}
	\begin{gathered}
		Y_{i, j, d} = \sigma(X_{i, j, d} + \sum_{h=1}^{H_r}    \sum_{w=1}^{W_r}   H_{in}  W_{in} (G^y)^T_{i, h}    (G^x)^T_{j, w} Y^r_{h, w, d})
		\\\forall  1 \leq i \leq H_{in}, 1 \leq j \leq W_{in}, 1 \leq d \leq D.
	\end{gathered}
	\label{eq:saliency sampler inv sampling}
\end{equation}

Instead of using bilinear sampling\cite{jaderberg2015spatial}, we compute the weights between the input and output pixels, and directly use the weighted sum as the output value. This approach can simplify the calculation, as it does not involve calculation of the coordinates. Furthermore, bilinear sampling may skip some pixels due to the possible non-uniform downsampling, but the proposed method takes all input pixels into account.

Note that the sampling function can be simply implemented by two batch matrix multiplications, which gives a complexity of $O(H_r H_{in} W_{in} D + H_r W_r W_{in} D)$ (when computed in y-axis first). The complexity is higher than bilinear sampling that has a complexity of $O(H_r W_r D)$. However, the weight matrices $G^y$ and $G^x$ contain at most $H_{in} + H_r$ and  $W_{in} + W_r$ non-zero elements
\footnote{
	Consider a weight matrix $G^y$ with dimensions $H_r \times H_{in}$. If the $(i, j)$-th element of the weight matrix is non-zero, the next non-zero index will be  $(i+\Delta i, j+\Delta j)$, where $\Delta i$, $\Delta j$ are non-negative integers with either $i+\Delta i > i$ or $j+\Delta j> j$. As a result, there are at most $H_{in} + H_r$  non-zero elements. The same  is true for $G^x$.
}
, respectively. If the sampling sizes scaled linearly with the input sizes, the complexity of the sampling function can be reduced to $O(H_{r} W_{r} D)$. Thus, an optimized implementation which considers the sparsity of the matrices could significantly reduce the complexity and latency, and allow SSBNet to scale well with high dimension input.

\section{Experiments}

\label{sec:experiments}

In this section, we first train SSBNet and the baseline models for image classification tasks on the ImageNet dataset\cite{ILSVRC15}, and then fine-tune the models to the object detection and instance segmentation tasks on the COCO dataset\cite{lin2014microsoft}. After that, we report the inference performance of SSBNet. All experiments were conducted with TensorFlow 2.6\cite{Abadi_TensorFlow_Large-scale_machine_2015} and Model Garden \cite{tensorflowmodelgarden2020}, and ran on TPU v2-8/v3-8 with bfloat16, except for Section \ref{subsec:latency}. For ease of presentation, we denote the configurations for the last $L$ groups of SSBNets by ($M_1$, ...,$M_L$). Here, $M_l$ indicates that the sampling size of the last $(L - l + 1)$-th group is $M_l \times M_l$.

\subsection{Image Classification}
\label{subsec:image classification}

For image classification, we trained SSBNets and the baseline models on the ImageNet\cite{ILSVRC15} dataset, which contains 1.28M training and 50k validation samples. We built SSBNets based on ResNet-D \cite{he2019bag}, ResNet-RS\cite{bello2021revisiting}, EfficientNet\cite{tan2019efficientnet} and BoTNet-S1 \cite{srinivas2021bottleneck}, and compare their performance with the original models. Due to limited resources, we only trained some variants of ResNet-RS and found that the results are close to the original work\cite{bello2021revisiting}. Thus, we will report other results directly from the original paper. For EfficientNet  and BotNet-S1, we were not able to reproduce the same results from the papers\cite{tan2019efficientnet, srinivas2021bottleneck}. For fair comparison, we trained and reported all variants that have similar complexity as SSB-EfficientNet and SSB-BoTNet-S1.

Note that in this paper, we focus on the theoretical improvement regarding the accuracy-FLOPS trade-off. In Section \ref{subsec:latency}, we will compare the inference time between SSBNet and the baselines, which shows that real speedup is achievable.

\begin{table}[t]
	\scriptsize
	\setlength{\tabcolsep}{2pt}
	\caption{ImageNet results of (SSB-)ResNet-D and (SSB-)ResNet-RS}
	\begin{tabular}[t]{@{}lc c c c c}
		\toprule
		Model & Input size& Params & FLOPS & Top-1(\%) \\
		\midrule[1pt]
		R-50 & $224\times224$ & 25.6M & 4.3G & $78.1$  \\
		S-50 & $224\times224$ & 25.6M & 3.0G &  $78.1$ \\
		\midrule
		R-101 & $224\times224$ & 44.6M & 8.0G & $79.5$ \\
		S-101 & $224\times224$ & 44.6M & 4.3G & $78.9$\\
		\midrule
		R-152 & $224\times224$ & 60.2M & 11.8G & $80.1$ \\
		S-152 & $224\times224$ & 60.3M & 5.6G & $79.2$ \\
		\midrule
		\multicolumn{5}{@{} l}{R:  ResNet-D\cite{he2019bag} \ \ S:  SSB-ResNet-D} \\
		\bottomrule
		\toprule
		Model & Input size & Params & FLOPS & Top-1(\%) \\
		\midrule[1pt]
		R-50 & $224 \times 224$ & 25.6M & 4.3G & $78.2$ \\
		S-50 & $224 \times 224$ & 25.6M & 3.0G & $78.2$ \\
		\midrule
		R-101 & $224 \times 224$ & 44.6M & 8.0G & $80.0$ \\
		S-101 & $224 \times 224$ & 44.6M & 4.3G & $79.5$ \\
		\midrule
		R-152 & $224 \times 224$ & 60.2M & 11.8G & $80.6$ \\
		S-152 & $224 \times 224$ & 60.3M & 5.6G & $80.1$ \\
		\midrule
		\multicolumn{5}{@{} l}{R:  ResNet-D\cite{he2019bag} + RandAugment\cite{cubuk2020randaugment}} \\
		\multicolumn{5}{@{} l}{S:  SSB-ResNet-D + RandAugment\cite{cubuk2020randaugment}} \\
		\bottomrule
	\end{tabular}
	\hfill
	\begin{tabular}[t]{@{}lc c c c c}
		\toprule
		Model & Input size& Params & FLOPS & Top-1(\%) \\
		\midrule[1pt]
		R-50 & $160\times160$ & 35.7M & 2.3G & $78.8$ \\
		S-50 & $160\times160$ & 35.7M & 1.6G &  $78.2$ \\
		\midrule
		R-101 & $160\times160$ & 63.6M & 4.2G & $80.3^*$ \\
		S-101 & $160\times160$ & 63.6M & 2.3G & $79.3$ \\
		\midrule
		R-101 & $192\times192$ & 63.6M & 6.0G & $81.3$ \\
		S-101 & $192\times192$ & 63.6M & 3.2G & $80.6$ \\
		\midrule
		R-152 & $192\times192$& 86.6M & 9.0G & $82.0^*$ \\
		S-152 & $192\times192$ & 86.7M & 4.3G & $81.0$ \\
		\midrule
		R-152 & $224\times224$ & 86.6M & 12.0G & $82.5$ \\
		S-152 & $224\times224$ & 86.7M & 5.8G & $81.7$ \\
		\midrule
		R-152 & $256\times256$ & 86.6M & 15.5G & $83.0^*$ \\
		S-152 & $256\times256$ & 86.7M & 7.5G & $82.1$ \\
		\midrule
		R-200 & $256\times256$ & 93.2M & 20.0G & $83.4^*$ \\
		S-200 & $256\times256$ & 93.3M & 9.4G & $82.6$ \\
		\midrule
		\multicolumn{5}{@{} l}{R:  ResNet-RS\cite{bello2021revisiting} \ \ S:  SSB-ResNet-RS} \\
		\multicolumn{5}{@{} l}{$^*$:  from the original paper} \\
		\bottomrule
	\end{tabular}
	\label{tab:imagenet_resnet_d_rs}
\end{table}

\subsubsection{Comparison with ResNet-D.} For ResNet-D \cite{he2019bag} and SSB-ResNet-D, we trained three scales with the configuration of ResNet50/101/152\cite{he2016deep}. We followed the training and testing settings of \cite{he2019bag} with batch size of 1024 and input size of $224\times224$. The sampling sizes of SSB-ResNet-D are $(16, 8, 4)$. The results are shown in the top-left table of Table \ref{tab:imagenet_resnet_d_rs}, which clearly demonstrate the advantage of SSB-ResNet-D. For example, SSB-ResNet-D-50 achieved similar accuracy as ResNet-D-50, where the FLOPS is reduced by 30\%. SSB-ResNet-D-101 has the same FLOPS as ResNet-D-50, but achieved 0.8\% higher performance.  The deepest SSB-ResNet-D-152 also performed only 0.3\% worse than ResNet-D-101, with 30\% less FLOPS.

In addition, we conducted experiments with RandAugment \cite{cubuk2020randaugment} as data augmentation. The number of transformations and the magnitude were 2 and 5, respectively. It can be observed from the bottom-left table of Table \ref{tab:imagenet_resnet_d_rs} that SSB-ResNet-D-101 outperformed ResNet-D-50 by 1.3\% accuracy with the same FLOPS, and SSB-ResNet-D-152 achieved similar accuracy as ResNet-D-101 but saved 30\% operations. This indicates that SSBNets can benefit more from stronger regularization.

\subsubsection{Comparison with ResNet-RS.} We also conducted experiments with ResNet-RS \cite{bello2021revisiting} by followed the same training settings in the original paper. We trained (SSB-)ResNet-RS-50/101/152/200, with different input sizes of $160\times160$/$192\times192$/$224\times224$/$256\times256$, and the sampling sizes are scaled to (12, 6, 3)/(14, 7, 3)/(16, 8, 4)/(18, 9, 5), respectively. The performance comparison between SSB-ResNet-RS and ResNet-RS is shown Table \ref{tab:imagenet_resnet_d_rs}(right). The SSB-ResNet-RS achieved competitive results. For example, the SSB-ResNet-RS-200 with input size of $256\times256$ achieved 0.6\% higher accuracy than ResNet-RS-152 with input size of $192\times192$, where the FLOPS is only 4.4\% higher. Figure \ref{fig:resnetssbresnet} compares SSB-ResNet-RS with ResNet-RS with less than 10 GFLOPS and shows that SSB-ResNet-RS achieved better accuracy to FLOPS ratio in different scales.

\begin{table}[t]
	\scriptsize
	\setlength{\tabcolsep}{2pt}
	\caption{ImageNet results of (SSB-)EfficientNet and (SSB-)BoTNet-S1}
	\begin{tabular}[t]{@{}lc c c c c}
		\toprule
		Model & Input size& Params & FLOPS & Top-1(\%) \\
		\midrule[1pt]
		E-B0 & $224\times224$ & 5.3M & 0.4G & $76.4$ \\
		S-B0 & $224\times224$ & 5.3M & 0.3G & $75.4$ \\
		\midrule
		E-B1 & $240\times240$ & 7.9M & 0.7G & $78.5$ \\
		S-B1 & $240\times240$ & 7.9M & 0.5G & $77.4$ \\
		\midrule
		E-B2 & $260\times260$ & 9.2M & 1.0G & $79.6$ \\
		S-B2 & $260\times260$ & 9.2M & 0.7G & $78.6$ \\
		\midrule
		E-B3 & $300\times300$ & 12.3M & 1.8G & $81.0$ \\
		S-B3 & $300\times300$ & 12.3M & 1.3G & $80.1$ \\
		\midrule
		\multicolumn{5}{@{} l}{E:  EfficientNet\cite{tan2019efficientnet} \ \ S:  SSB-EfficientNet} \\
		\bottomrule
	\end{tabular}
	\hfill
	\begin{tabular}[t]{@{}lc c c c c}
		\toprule
		Model & Input size& Params & FLOPS & Top-1(\%) \\
		\midrule[1pt]
		B-59 & $224\times224$ & 30.5M & 7.3G & $81.1$ \\
		S-59 & $224\times224$ & 30.5M & 3.8G & $80.3$ \\
		\midrule
		B-110 & $224\times224$ & 51.7M & 10.9G & $82.1$ \\
		S-110 & $224\times224$ & 51.8M & 5.1G & $81.2$ \\
		\midrule
		B-128 & $256\times256$ & 69.1M & 19.3G & $82.9$ \\
		S-128 & $256\times256$ & 69.1M & 8.1G & $82.0$ \\
		\midrule
		B-77 & $320\times320$ & 47.9M & 23.3G & $-$ \\
		S-77 & $320\times320$ & 47.9M & 10.2G & $82.5$ \\
		\midrule
		\multicolumn{5}{@{} l}{B:  BoTNet-S1\cite{srinivas2021bottleneck} \ \ S:  SSB-BoTNet-S1} \\
		\bottomrule
	\end{tabular}
	\label{tab:imagenet_efficientnet_botnet}
\end{table}

\subsubsection{Comparison with EfficientNet.} For EifficientNet\cite{tan2019efficientnet}, we followed the original setting, except that we used a batch size of 1024. Due to the use of $5\times5$ convolutions, we applied larger sampling size for EfficientNet. Specifically, we used (20, 10, 10, 5, 5)/(22, 11, 11, 5, 5)/(24, 12 ,12 ,6, 6) and (26, 13 ,13 ,7, 7) for SSB-EfficientNet-B0/1/2/3, respectively.

However, adaptive sampling did not improve EfficientNet in our experiments as shown in Table \ref{tab:imagenet_efficientnet_botnet}(left). For example, SSB-EfficientNet-B2 has less number of operations than  EfficientNet-B2, but nearly the same number of operations and accuracy as  EfficientNet-B1. This may be due to the fact that EfficientNet is designed by neural architecture search, thus the network configurations and training parameters do not transfer well to SSB-EfficientNet. We also note that the speed-up in EfficientNet is limited when compared with ResNet. This is because EfficientNet has more groups of layers and we didn't replace the first layer in each group, due to the reason discussed in Section \ref{subsec:ssbnet}.

\subsubsection{Comparison with BoTNet-S1.}
The recent development of Visual Transformers\cite{dosovitskiy2021an} has gained attention in the research community. However, training Transformers is challenging. For example, larger dataset or additional augmentation is required\cite{dosovitskiy2021an, touvron2021training}. To evaluate the compatibility of adaptive sampling with self-attention layer, we built SSBNet from BoTNet-S1\cite{srinivas2021bottleneck}, which is a hybrid network composed of both convolutional and self-attention layers. It inherited the techniques from modern CNNs, including the choice of normalization layers, optimizers, and training setting.

Results in Table \ref{tab:imagenet_efficientnet_botnet}(right) show that SSB-BoTNet-S1 achieved better accuracy to FLOPS trade-off. For example, SSB-BoTNet-S1-110 performed similar as BoTNet-S1-59, but with 30\% less FLOPS; SSB-BoTNet-S1-77 achieved 0.4\% higher accuracy than BoTNet-S1-110, but 6\% less FLOPS. The results also suggest that adaptive sampling does not only improve CNNs, but also hybrid networks that utilize self-attention.  Figure \ref{fig:botnetssbbotnet} shows the comparison between SSB-BoTNet-S1 and BoTNet-S1 that have less than 11 GFLOPS, where SSB-BoTNet-S1 achieved better accuracy to FLOPS ratio.

\begin{figure}[t]
	\begin{subfigure}[t]{0.465\textwidth}
		\includegraphics[trim={0 22.3cm 0 0},clip,width=1.0\linewidth]{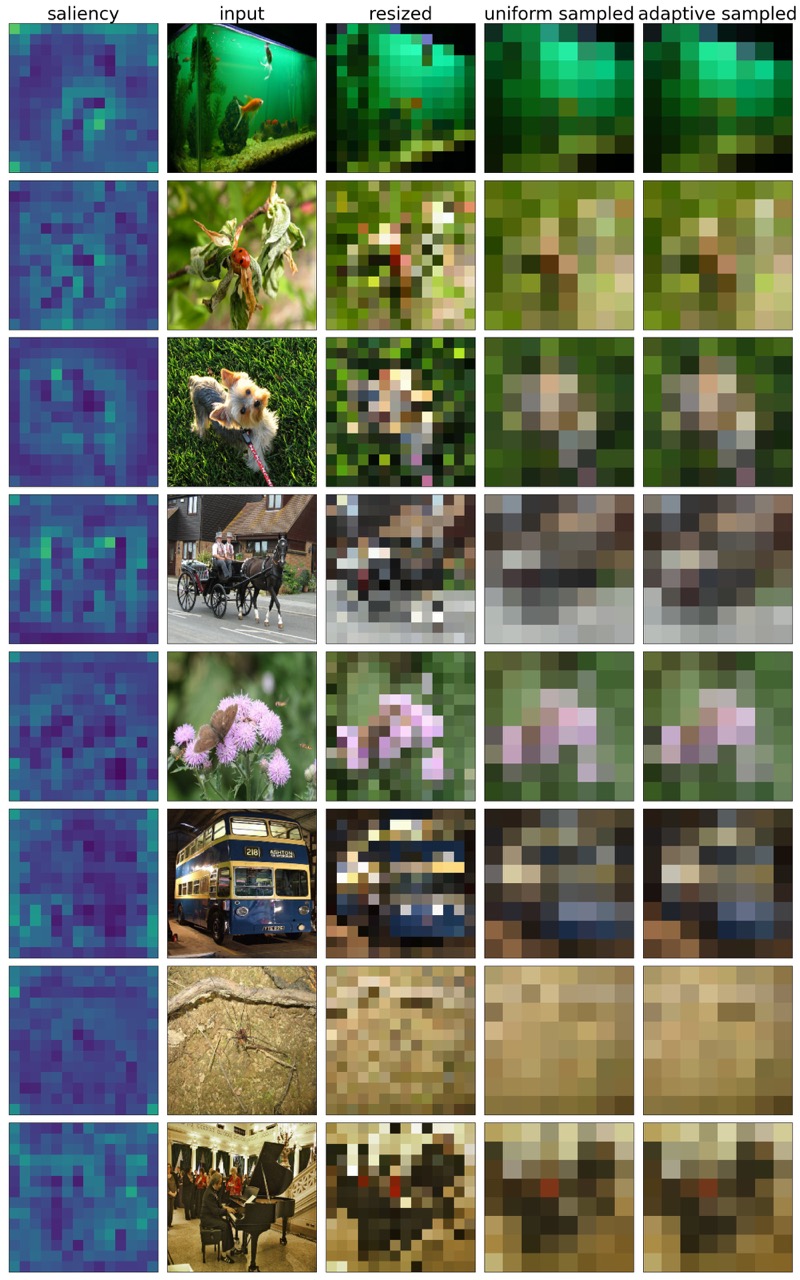}
		\caption{Example 1}
		\label{fig:vis1}
	\end{subfigure}
	\hfill
	\begin{subfigure}[t]{0.465\textwidth}
		\includegraphics[trim={0 22.3cm 0 0},clip,width=1.0\linewidth]{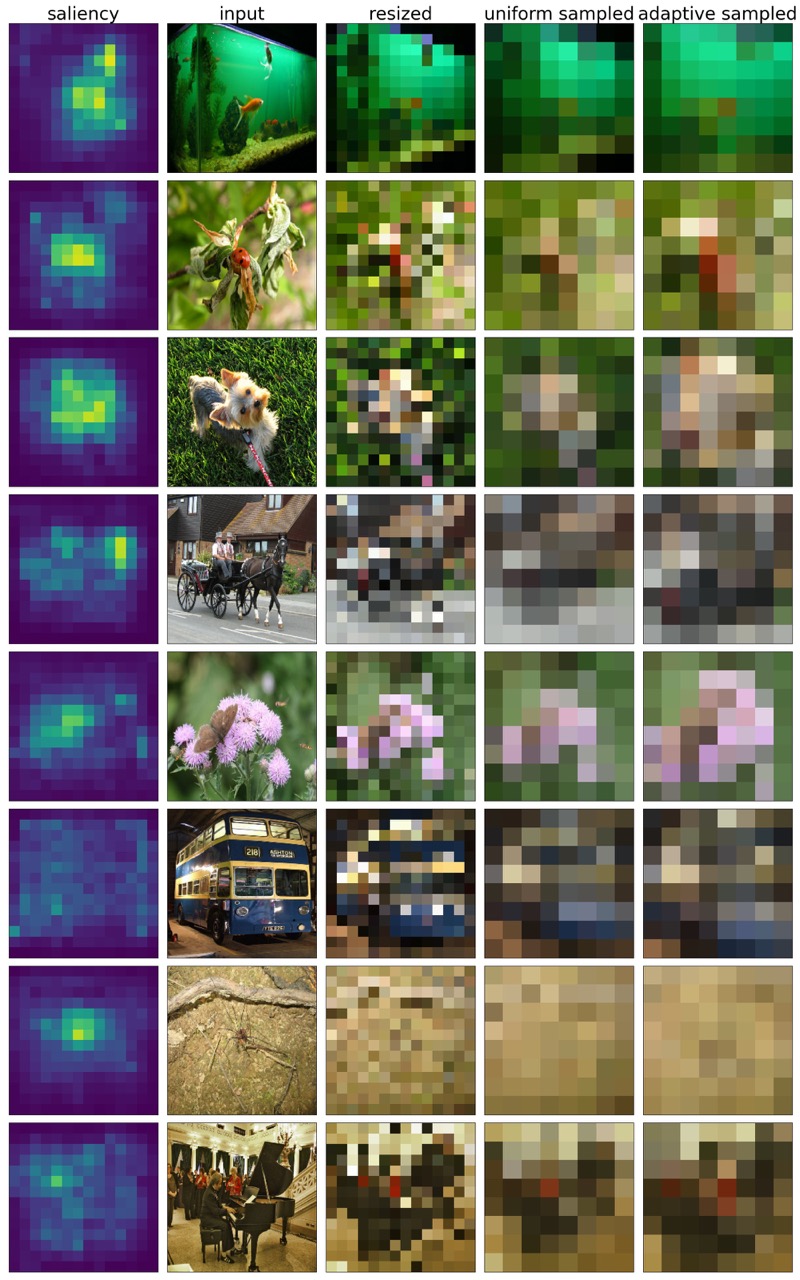}
		\caption{Example 2}
		\label{fig:vis2}
	\end{subfigure}
	\caption{Examples of the saliency map and sampling output. (a) and (b) are the samples from different layers of the SSB-ResNet-RS-152.}
	\label{fig:visx2}
\end{figure}

\subsubsection{Visualization.} 
The outputs of two sampled layers of SSBNet are shown in Figure \ref{fig:visx2}. While the high dimension features are hard to visualize, we first resized the original images to the same size as the sampler input, and then applied sampling on the resized images for visualization. Figure \ref{fig:vis1} shows that the first reported layer samples from the whole image, where the background is weighted heavier; Figure \ref{fig:vis2} shows that the second reported layer is able to zoom into smaller regions when performing downsampling, which can better preserve the discriminative features in these regions. The results also suggest that different layers of the SSBNet zoom into different regions, which justifies the use of adaptive sampling in multiple layers.

\subsection{Object Detection and Instance Segmentation}
\label{subsec:detection and segmentation}

\begin{table}[t]
	\centering
	\scriptsize
	\setlength{\tabcolsep}{2pt}
	\caption{COCO-2017\cite{lin2014microsoft} results of SSB-ResNet-D/ResNet-D\cite{he2019bag} with FPN\cite{lin2017feature} and Mask R-CNN\cite{he2017mask}}
	\begin{tabular}[t]{@{}lc c c c}
		\toprule
		& Model & AP\textsubscript{box} &  AP\textsubscript{mask} \\
		\midrule[1pt]
		12 Epochs & ResNet-D-50 & 40.14&  35.72&  \\
		& SSB-ResNet-D-50 & 40.67 & 35.98 \\
		\midrule
		24 Epochs & ResNet-D-50 & 41.95&  37.17&  \\
		& SSB-ResNet-D-50 & 42.36 & 37.41 \\
		\bottomrule
	\end{tabular}
	\begin{tabular}[t]{@{}lc c c c}
		\toprule
		& Model & AP\textsubscript{box} &  AP\textsubscript{mask} \\
		\midrule[1pt]
		36 Epochs & ResNet-D-50 & 42.50 &  37.56 &  \\
		& SSB-ResNet-D-50 & 42.89 & 37.84 \\
		\bottomrule
	\end{tabular}
	\label{tab:coco result}
\end{table}

We also evaluated the performance of SSBNet for object detection and instance segmentation tasks on the COCO-2017 dataset\cite{lin2014microsoft} which contains 118K images in the training set and 5K images for validation. For that purpose, we used the pre-trained ResNet-D-50\cite{he2019bag} and SSB-ResNet-D-50 with FPN\cite{lin2017feature} as the backbone, and applied Mask R-CNN\cite{he2017mask} for object detection and segmentation. The same was applied to the baseline models for comparison purposes. We used the default setting of Mask R-CNN in Model Garden\cite{tensorflowmodelgarden2020}, with input size of $1024\times1024$, batch size of 64 and a learning rate of 0.01. Horizontal flipping and Scale jitter with a range between 0.8 and 1.25 were applied. For SSB-ResNet-D, we used the sampling sizes of (72, 36, 18) for the last 3 groups. The results of training with 12/24/36 epochs are reported.

The results in Table \ref{tab:coco result} show that SSBNet is transferable to new tasks with high performance. While the pre-trained ResNet-D-50 and SSB-ResNet-D-50 have similar accuracy on ImageNet\cite{ILSVRC15 }, the SSB-ResNet-D-50 performs slightly better than ResNet-D-50 on COCO-2017, with less operations. This may be due to two reasons: 1.) there are paddings to the images, but SSBNet can zoom into the non-padding regions such that no computation is wasted, 2.) the images from COCO dataset have higher resolution than those from ImageNet.

\subsection{Latency Comparison between SSBNet and the Original Networks}
\label{subsec:latency}
\begin{table}[t]
	\scriptsize
	\setlength{\tabcolsep}{2pt}
	\caption{Latency comparison between (SSB-)ResNet-RS and (SSB-)BoTNet-S1}
	\begin{tabular}[t]{@{} c c c c c}
		\toprule
		Model & Input size & Params & FLOPS & Latency(ms) \\
		\midrule[1pt]
		R-50 & $160 \times 160$ & 35.7M & 2.3G & $130/163$ \\
		S-50 & $160 \times 160$ & 35.7M & 1.6G &  $117/138$ \\
		\midrule
		R-101 & $192 \times 192$ & 63.6M & 6.0G & $298/381$ \\
		S-101 & $192 \times 192$ & 63.6M & 3.2G & $235/267$ \\
		\midrule
		R-152 & $224 \times 224$ & 86.6M & 12.0G & $565/726$ \\
		S-152 & $224 \times 224$ & 86.7M & 5.8G & $427/472$ \\
		\midrule
		R-200 & $256 \times 256$ & 93.2M & 20.0G & $988/1244$ \\
		S-200 & $256 \times 256$ & 93.2M & 9.4G & $744/822$ \\
		\midrule
		\multicolumn{5}{@{} l}{R: ResNet-RS\cite{bello2021revisiting} \ \ S:  SSB-ResNet-RS} \\
		\multicolumn{5}{@{} l}{First number: Latency on V100 GPU} \\
		\multicolumn{5}{@{} l}{Second number: Latency on 3090 GPU} \\
		\bottomrule
	\end{tabular}
	\hfill
	\begin{tabular}[t]{@{} c c c c c}
		\toprule
		Model & Input size & Params & FLOPS & Latency(ms) \\
		\midrule[1pt]
		B-59 & $224 \times 224$ & 30.5M & 7.3G & $404/469$ \\
		S-59 & $224 \times 224$ & 30.5M & 3.8G & $291/301$ \\
		\midrule
		B-110 & $224 \times 224$ & 51.7M & 10.9G & $559/674$ \\
		S-110 & $224 \times 224$ & 51.8M & 5.1G & $386/406$ \\
		\midrule
		B-128 & $256 \times 256$ & 69.1M & 19.3G & $925/1127$ \\
		S-128 & $256 \times 256$ & 69.1M & 8.1G & $612/639$ \\
		\midrule
		B-77 & $320 \times 320$ & 47.9M & 23.3G & $1234/1487$ \\
		S-77 & $320 \times 320$ & 47.9M & 10.2G & $773/784$ \\
		\midrule
		\multicolumn{5}{@{} l}{B: BoTNet-S1\cite{srinivas2021bottleneck} \ \ S:  SSB-BoTNet-S1} \\
		\multicolumn{5}{@{} l}{First number: Latency on V100 GPU} \\
		\multicolumn{5}{@{} l}{Second number: Latency on 3090 GPU} \\
		\bottomrule
	\end{tabular}
	\label{tab:latency}
\end{table}

To explore the actual speed-up by adaptive sampling, we implemented the sampling function in TensorFlow 2.6\cite{Abadi_TensorFlow_Large-scale_machine_2015} and CUDA \cite{cuda}, and performed comparison between ResNet-RS \cite{bello2021revisiting}, SSB-ResNet-RS, BoTNet-S1 \cite{srinivas2021bottleneck}, and SSB-BoTNet-S1. Experiments were conducted on V100 and 3090 GPU with float16. We report the results from two GPUs as we noticed the difference in performance. Specifically, V100 is commonly used in the literatures, but our implementation performs better in 3090, which is possibly due to the degree of optimization. The results are reported by the batch latency with size of 1024.

The results in Table \ref{tab:latency} show that actual speed-up is achievable with adaptive sampling. For example, the latency of SSB-ResNet-RS-200 is reduced by up to 34\% when compared with ResNet-RS-200, where ideally the latency can be reduced by 53\%. The latency of SSB-BoTNet-S1-77 is 47\% lower than BoTNet-S1-77, which is close to the theoretical improvement, i.e. 56\%. We would like to highlight that we only did limited implementation optimization over SSBNet. We expect further improvement with better optimization. 

\section{Ablation study}
In this section, we provide results of additional experiments to justify the use of adaptive sampling in deep neural networks.

\subsection{Comparison between Uniform and Adaptive Sampling}
\label{subsec:uvsa}

\begin{figure}[t]
	\begin{subfigure}[t]{0.475\textwidth}
		\centering
 		\includegraphics[width=4.5cm,trim={0 0 0 1.2cm 0},clip]{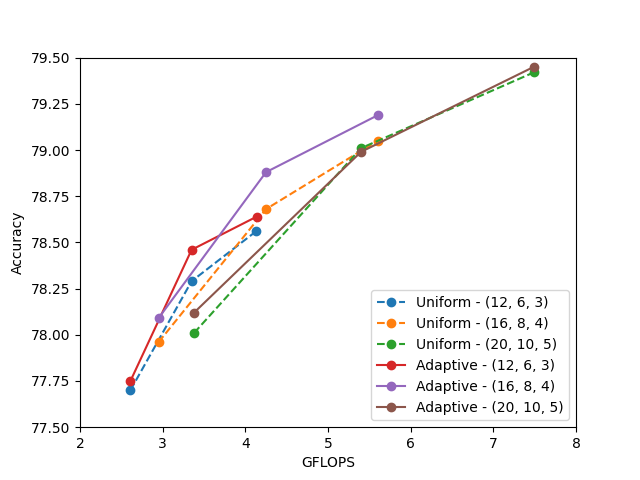}
		\caption{Normal}
		\label{fig:uvsa}
	\end{subfigure}
	\hfill
	\begin{subfigure}[t]{0.475\textwidth}
		\centering
		\includegraphics[width=4.5cm,trim={0 0 0 1.2cm 0},clip]{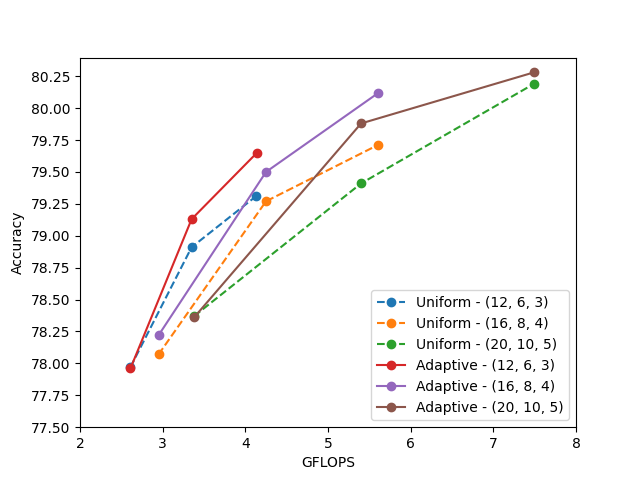}
		\caption{With RandAugment \cite{cubuk2020randaugment}}
		\label{fig:uvsarandaug}
	\end{subfigure}
	\caption{Comparison between uniform and adaptive sampling. In each line, the three points denote SSB-ResNet-D-50/101/152, respectively.}
	\label{fig:main}
\end{figure}

To validate the effectiveness of adaptive sampling utilized in SSBNet, we compared the performance of two SSB-ResNet-D networks, which applied adaptive and uniform sampling, respectively. For a fair comparison, the two networks used the same sampling mechanism (Equation \ref{eq:saliency sampler sampling}).

The results in Figure \ref{fig:uvsa} show that, in general, the networks with adaptive sampling outperform the networks that applied uniform sampling. We observed the largest difference from SSB-ResNet101-D, where the adaptive model achieved 0.2\% higher accuracy. In Figure \ref{fig:uvsarandaug}, results with RandAugment\cite{cubuk2020randaugment} are shown, where models with adaptive sampling obtain larger improvement than those with uniform sampling. For example, at sampling sizes (16, 8, 4), the SSB-ResNet152-D with adaptive sampling achieved 0.4\% higher accuracy. 

The results suggest that adaptive sampling is a better choice for downsampling in SSBNet. In addition, the results show that the SSB-ResNet-D with sampling sizes of (16, 8, 4) achieved a good trade-off between accuracy and FLOPS at different depths. Thus, we used this configuration as default.

\subsection{Comparison with Other Sampling Methods}
\label{sec:addition_other_sampling}

\begin{figure}[t]
	\begin{subfigure}[t]{0.475\textwidth}
		\centering
		\includegraphics[width=4.5cm,trim={0 0 0 1.2cm 0},clip]{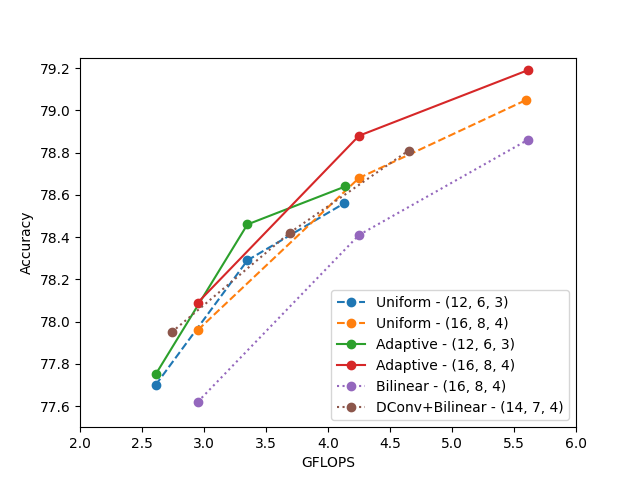}
		\caption{Normal}
		\label{fig:uvsavsothers}
	\end{subfigure}
	\begin{subfigure}[t]{0.475\textwidth}
		\centering
		\includegraphics[width=4.5cm,trim={0 0 0 1.2cm 0},clip]{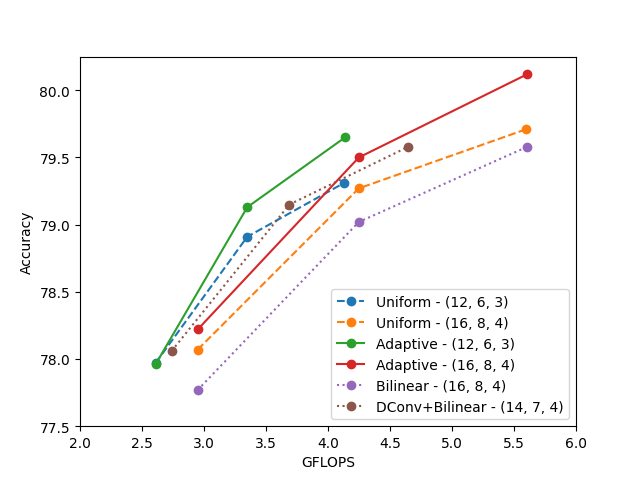}
		\caption{With RandAugment \cite{cubuk2020randaugment}}
		\label{fig:uvsavsothersrandaug}
	\end{subfigure}
	\caption{Comparison between adaptive sampling and other sampling methods. In each line, the three points denote SSB-ResNet50/101/152-D, respectively.}
	\label{fig:uvsavsothers2}
\end{figure}

To compare different down/upsampling mechanisms, we conducted experiments with different sampling methods that can be applied in the bottleneck structure: 1) the proposed adaptive sampling; 2) the uniform sampling used in Section \ref{subsec:uvsa}; 3) the uniform sampling with bilinear interpolation; and 4) the depthwise convolution for downsampling with bilinear sampling for upsampling  \cite{li2019hbonet}. For bilinear sampling, we used sampling sizes of $(16, 8, 4)$ as it is the common choice in our paper. For 4), the kernel size and stride of depthwise convolutions are $5$ and $2$ respectively, with sampling sizes of $(14, 7, 4)$. For a fair comparison, we included the results of adaptive sampling with sizes $(16, 8, 4)$ and  $(12, 6, 3)$, such that the cost of 4) falls between them. 

Figure \ref{fig:uvsavsothers} shows the results with basic training setting and Figure \ref{fig:uvsavsothersrandaug} shows the results with RandAugment \cite{cubuk2020randaugment}. In both settings, adaptive sampling outperformed other methods, especially when the model is deeper and additional data augmentation is used. Surprisingly, although method 2) is also a uniform sampling method, it outperformed the widely utilized method 3).

\section{Conclusion}
In this paper, we proposed a novel architecture to apply adaptive sampling in the main building block of deep neural networks. The proposed SSBNet outperformed other benchmarks in both image classification and object detection tasks. Different from most existing works that applied adaptive sampling for specific tasks\cite{jaderberg2015spatial, recasens2018learning, zheng2019looking, jin2022learning, thavamani2021fovea, dai2017deformable} and performed very few sampling operations, the proposed structure can work as a backbone network and be transferred to different tasks. Visualization illustrated SSBNet's capability  in sampling different regions at different layers and ablation studies demonstrated that adaptive sampling is more efficient than uniform sampling.

The results in this paper suggest that adaptive sampling is a promising mechanism in deep neural networks. We expect that designing the network with adaptive sampling from scratch and fine-tuning the training process may provide further performance improvement.

\section*{Acknowledgement}
This work was supported by the Cloud TPUs from Google's TPU Research Cloud (TRC) and the HKUST-WeBank Joint Lab under Grant WEB19EG01-L. 

%
%
\bibliographystyle{splncs04}
\bibliography{egbib}
\clearpage

\appendix
\section*{Appendix}
\section{Additional Results with Different Saliency Networks}
Besides the $1\times1$ convolutional layer utilized for the saliency map estimation in the main paper, we also conducted additional experiments by increasing the kernel sizes of the convolutional layers from $1\times1$ to $3\times3$ and $5\times5$. The experiments were performed with ResNet-D-50/101/152 \cite{he2016deep, he2019bag}. However, we did not notice any performance difference. This may be due to the fact that the saliency networks take the deep features as their inputs, which have large effective kernel sizes \cite{luo2016understanding}. As a result, increasing the kernel sizes in the saliency networks did not lead to any performance improvement.

\section{Comparison with Other Adaptive Networks}

\begin{table}[t]
	\centering
	\scriptsize
	\setlength{\tabcolsep}{4pt}
	\caption{Comparison to stochastic sampling-interpolation networks}
	\begin{tabular}[t]{@{}lc c c c}
		\toprule
		& Model & FLOPS & Top-1(\%) \\
		\midrule[1pt]
		120 Epochs & Baseline & 4.3G &  78.1  \\
		& SSB & 3.0G & 78.1 \\
		& SSIN, $\lambda$=0.001 & 3.9G &  76.2 \\
		& SSIN, $\lambda$=0.005 & 3.4G &  75.6 \\
		& SSIN, $\lambda$=0.010 & 3.2G &  75.8 \\
		& SSIN, $\lambda$=0.015 & 3.0G &  75.3 \\
		\midrule[1pt]
		200 Epochs & SSIN, $\lambda$=0.001 & 4.1G & 77.4 \\
		& SSIN, $\lambda$=0.005 & 3.7G & 76.6 \\
		& SSIN, $\lambda$=0.010 & 3.3G & 75.5 \\
		& SSIN, $\lambda$=0.015 & 3.0G & 73.3 \\
		\midrule
		\multicolumn{4}{@{} l}{Baseline:  ResNet-D-50\cite{he2019bag}} \\
		\multicolumn{4}{@{} l}{SSB:  SSB-ResNet-D-50} \\
		\multicolumn{4}{@{} l}{SSINI:  SSINI\cite{xie2020spatially} + ResNet-D-50\cite{he2019bag}} \\
		\bottomrule
	\end{tabular}
	\label{tab:ssb_vs_ssi}
\end{table}

In the main paper, we referred to adaptive sampling as the methods that perform geometric samplings or transformations on the images or feature maps, where the samplings or transformations depend on the inputs. Examples of adaptive sampling methods include spatial transformer \cite{jaderberg2015spatial} and saliency sampler \cite{recasens2018learning}. There are also approaches that adaptively skip the inference on individual units, such as pixels or blocks, to reduce the latency. For example, the stochastic sampling-interpolation network \cite{xie2020spatially} samples pixels for inference, where the SBNet \cite{ren2018sbnet} selects blocks.

In table \ref{tab:ssb_vs_ssi}, we provided preliminary comparison results between SSBNet and the stochastic sampling-interpolation network \cite{xie2020spatially} (referred to as SSIN for abbreviation), a recently proposed adaptive inference network. Following the same setting as the main paper, we trained SSBNet and SSIN based on ResNet-D-50\cite{he2019bag} on ImageNet\cite{ILSVRC15}. For SSIN, instead of the standard 120 epochs of training, we trained the models with 200 epochs, following the original paper which used a longer training for the adaptive networks. We reported results of SSIN with loss weight $\lambda$ of \{0.001, 0.005, 0.010, 0.015\}.

In our experiments, SSB-ResNet-D-50 achieved better accuracy than SSIN. Notice that in the SSIN paper, the authors trained ResNet34 for their experiments, where we utilized ResNet-D-50, which is a deeper and improved version of ResNet\cite{he2016deep}. The performance drop of SSIN reported here indicates that it may not fit the more complicated network and training scheme used in this paper. Compared to the baseline, the SSB-ResNet-D-50 has no drop in accuracy but with 30\% less FLOPS.

\section{Detailed Results of Ablation Studies}
In Tables \ref{tab:imagenet_sampling} and \ref{tab:imagenet_sampling_ra},
more detailed results regarding the computation complexity, i.e., FLOPS, and accuracy of models with different sampling methods and sampling sizes are shown. All results are reported by the average of 3 runs of SSB-ResNet-D, which is based on ResNet-D\cite{he2019bag}. As mentioned in the paper, four sampling methods are tested: 1) the proposed adaptive sampling in SSBNet, 2) the uniform sampling with the sampling mechanism in SSBNet (Equation 10 in the paper), 3) the uniform sampling with bilinear interpolation, and 4) the depthwise convolution for downsampling with bilinear interpolation for upsampling\cite{li2019hbonet}.

The results show that the models with adaptive sampling outperform the ones with uniform sampling on average, where the differences are larger with RandAugment\cite{cubuk2020randaugment}. With RandAugment, the SSB-ResNet-D-152 with adaptive sampling and sampling sizes of (16, 8, 4) outperformed all other networks that have similar complexity by at least 0.4\% in accuracy; the SSB-ResNet-D-152 with adaptive sampling and sampling sizes of (12, 6, 3) outperformed the network with depthwise convolution and bilinear sampling in accuracy, but with 11\% less computation.

\section{More results on Visualization }
\begin{figure}[t]
	\begin{subfigure}[t]{0.475\textwidth}
		\includegraphics[trim={0 5.6cm 0 0},clip,width=1.0\linewidth]{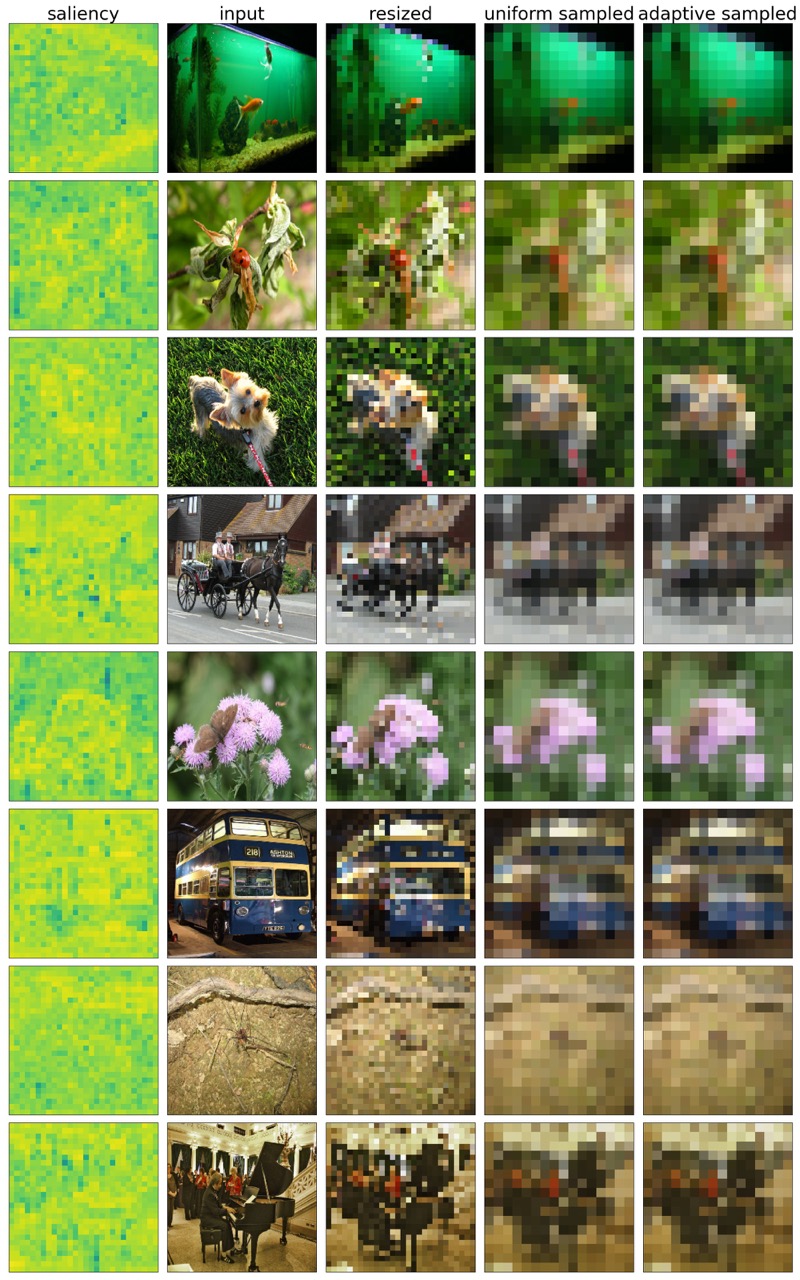}
		\caption{Example 1}
		\label{fig:visx2_1a}
	\end{subfigure}
	\hfill
	\begin{subfigure}[t]{0.475\textwidth}
		\includegraphics[trim={0 5.6cm 0 0},clip,width=1.0\linewidth]{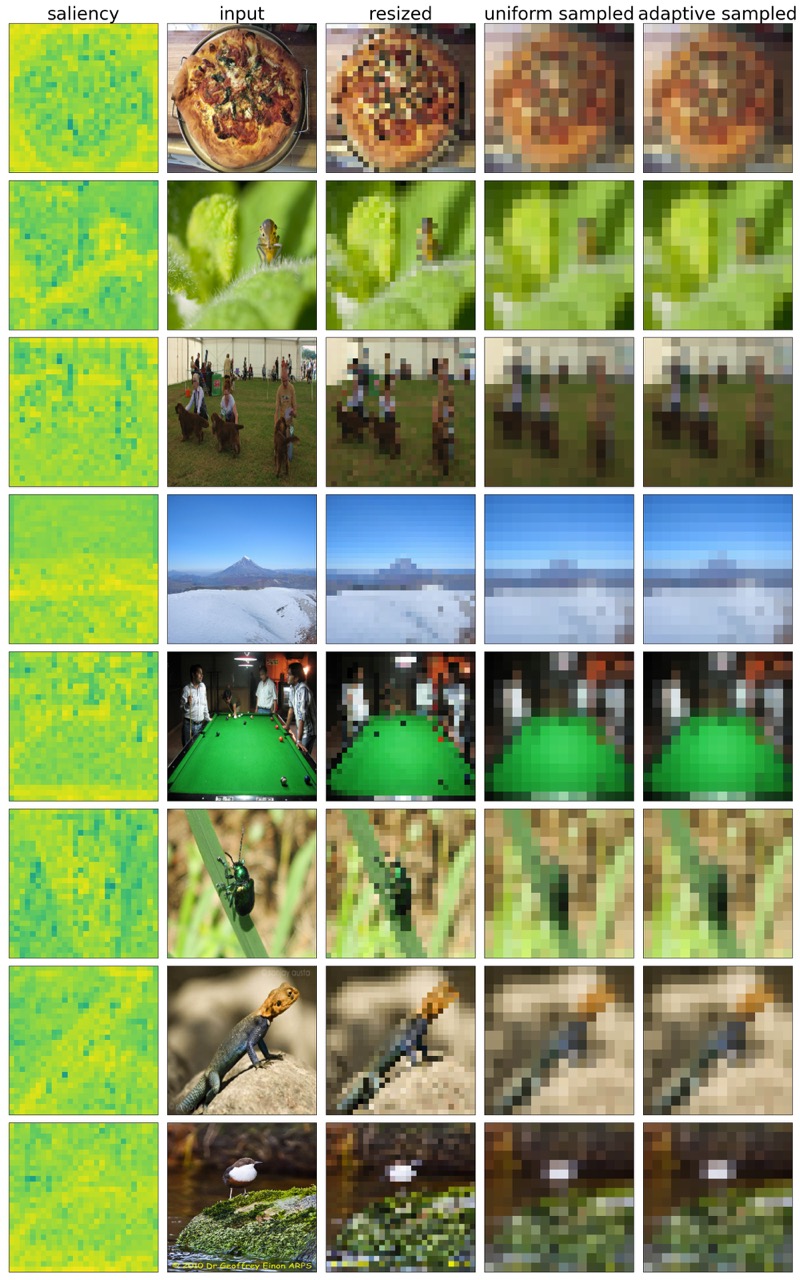}
		\caption{Example 2}
		\label{fig:visx2_1b}
	\end{subfigure}
	\caption{Layer 2-2}
	\label{fig:visx2_1}
\end{figure}

\begin{figure}[t]
	\begin{subfigure}[t]{0.475\textwidth}
		\includegraphics[trim={0 5.6cm 0 0},clip,width=1.0\linewidth]{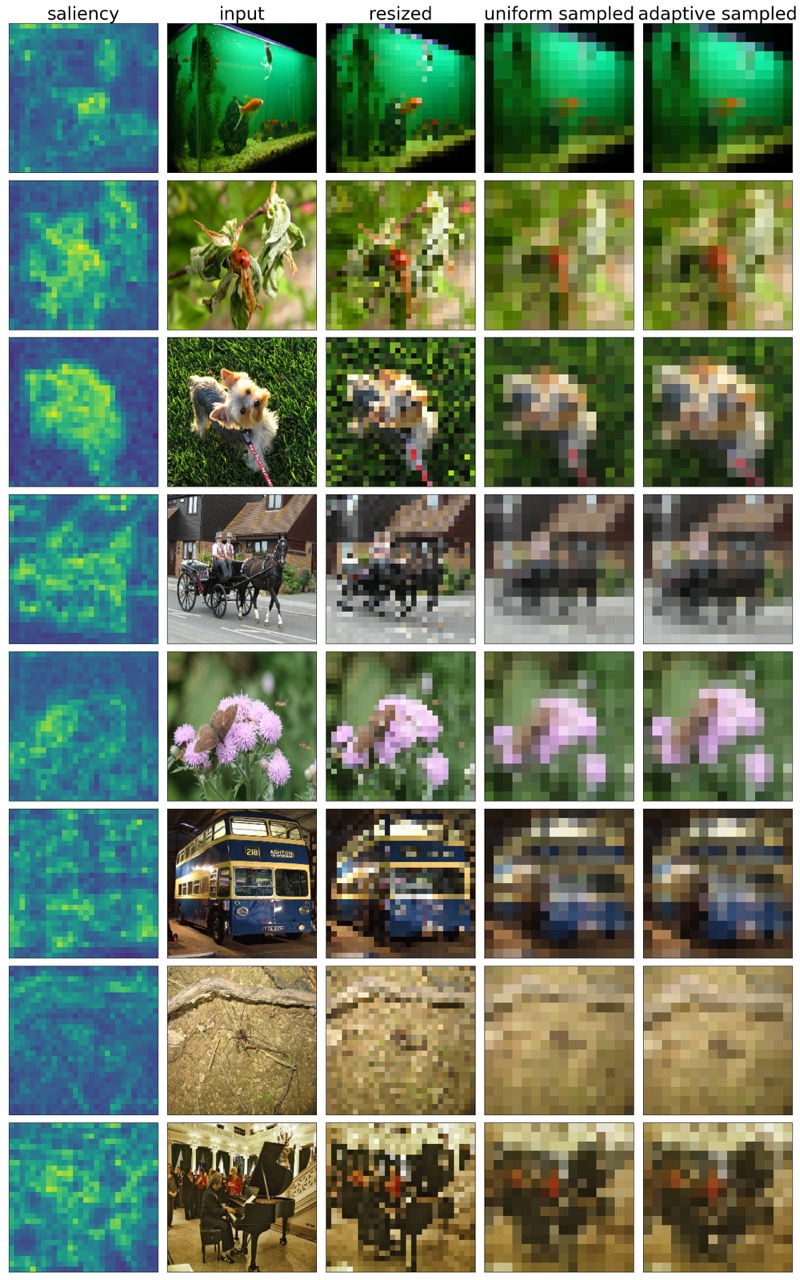}
		\caption{Example 1}
		\label{fig:visx2_2a}
	\end{subfigure}
	\hfill
	\begin{subfigure}[t]{0.475\textwidth}
		\includegraphics[trim={0 5.6cm 0 0},clip,width=1.0\linewidth]{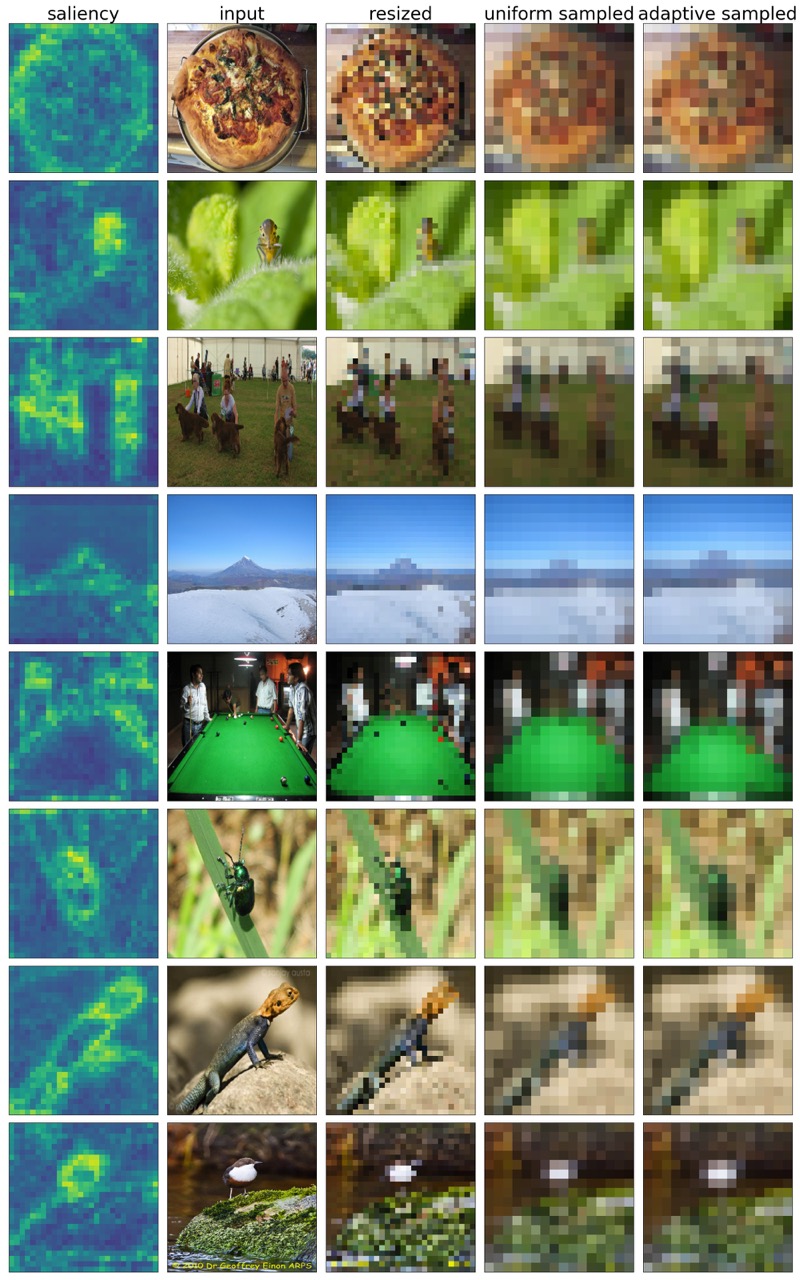}
		\caption{Example 2}
		\label{fig:visx2_2b}
	\end{subfigure}
	\caption{Layer 2-7}
	\label{fig:visx2_2}
\end{figure}

\begin{figure}[t]
	\begin{subfigure}[t]{0.475\textwidth}
		\includegraphics[trim={0 5.6cm 0 0},clip,width=1.0\linewidth]{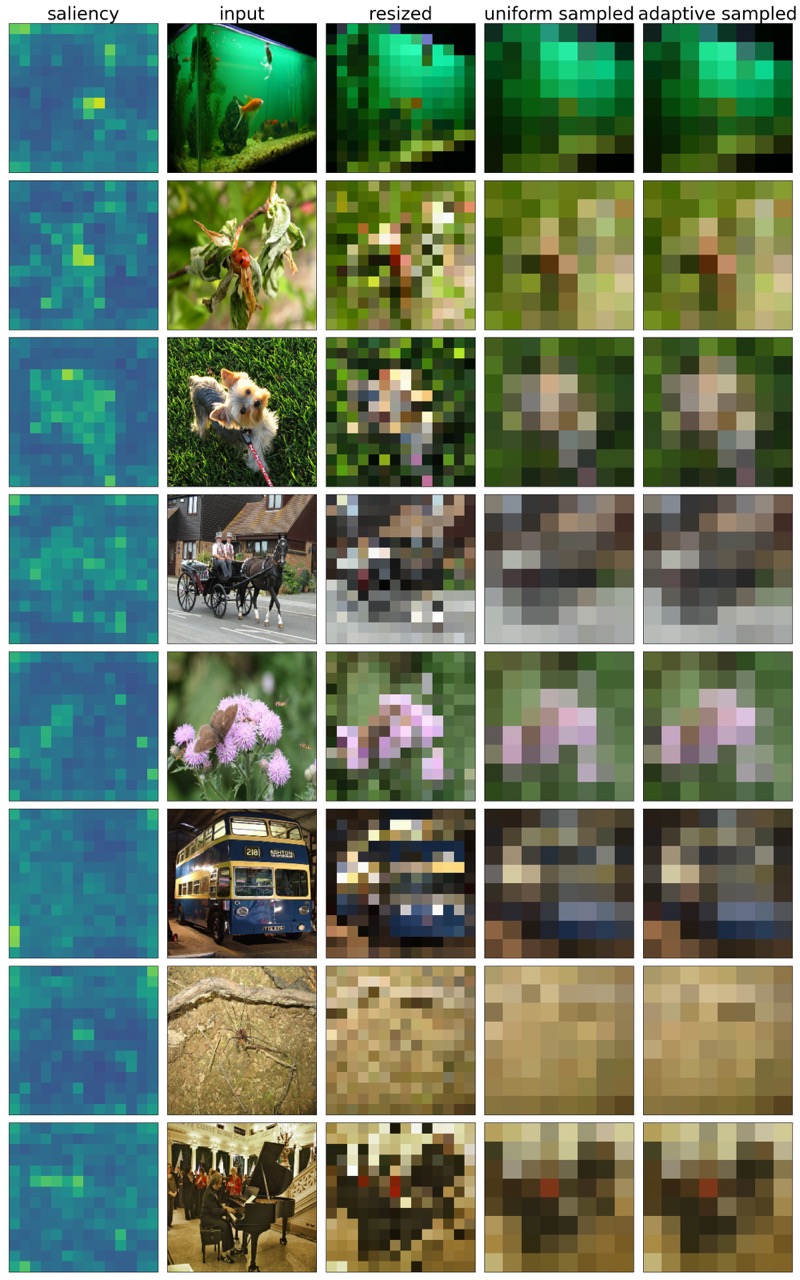}
		\caption{Example 1}
		\label{fig:visx2_3a}
	\end{subfigure}
	\hfill
	\begin{subfigure}[t]{0.475\textwidth}
		\includegraphics[trim={0 5.6cm 0 0},clip,width=1.0\linewidth]{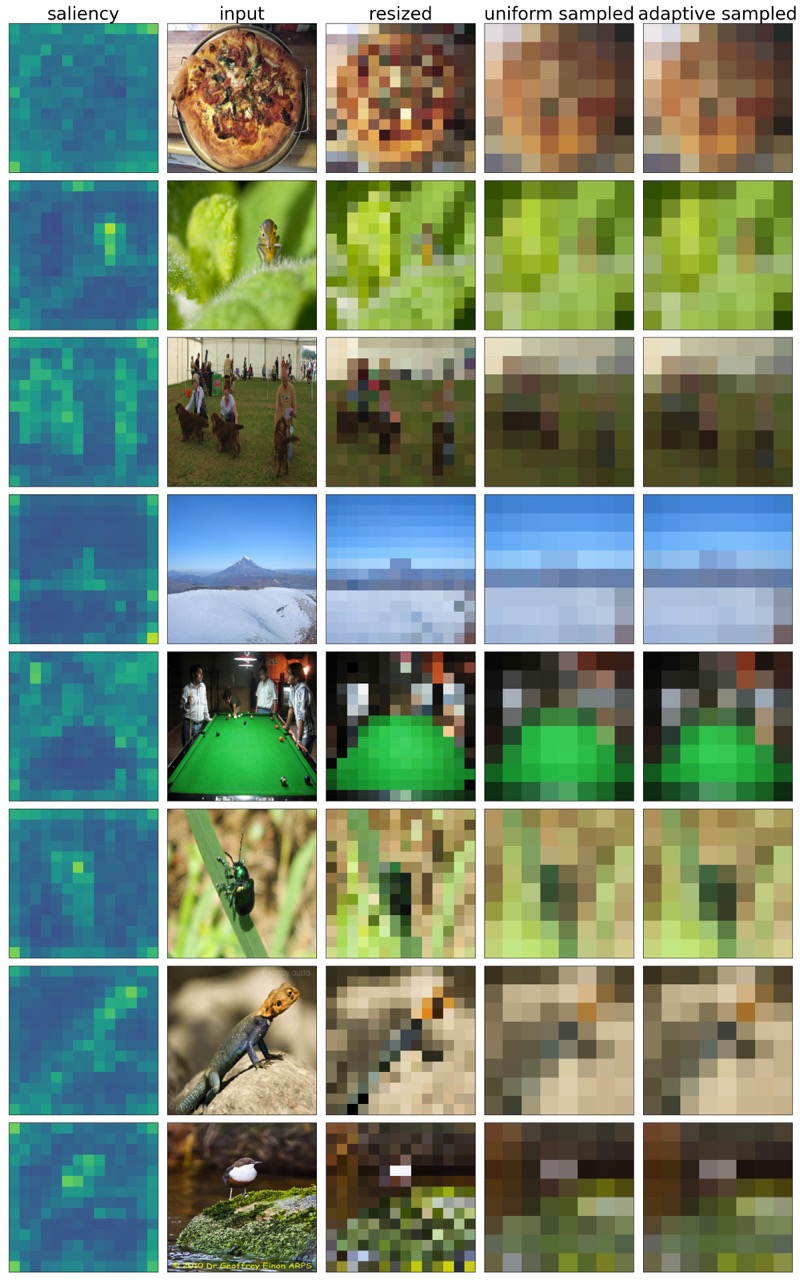}
		\caption{Example 2}
		\label{fig:visx2_3b}
	\end{subfigure}
	\caption{Layer 3-5}
	\label{fig:visx2_3}
\end{figure}

\begin{figure}[t]
	\begin{subfigure}[t]{0.475\textwidth}
		\includegraphics[trim={0 5.6cm 0 0},clip,width=1.0\linewidth]{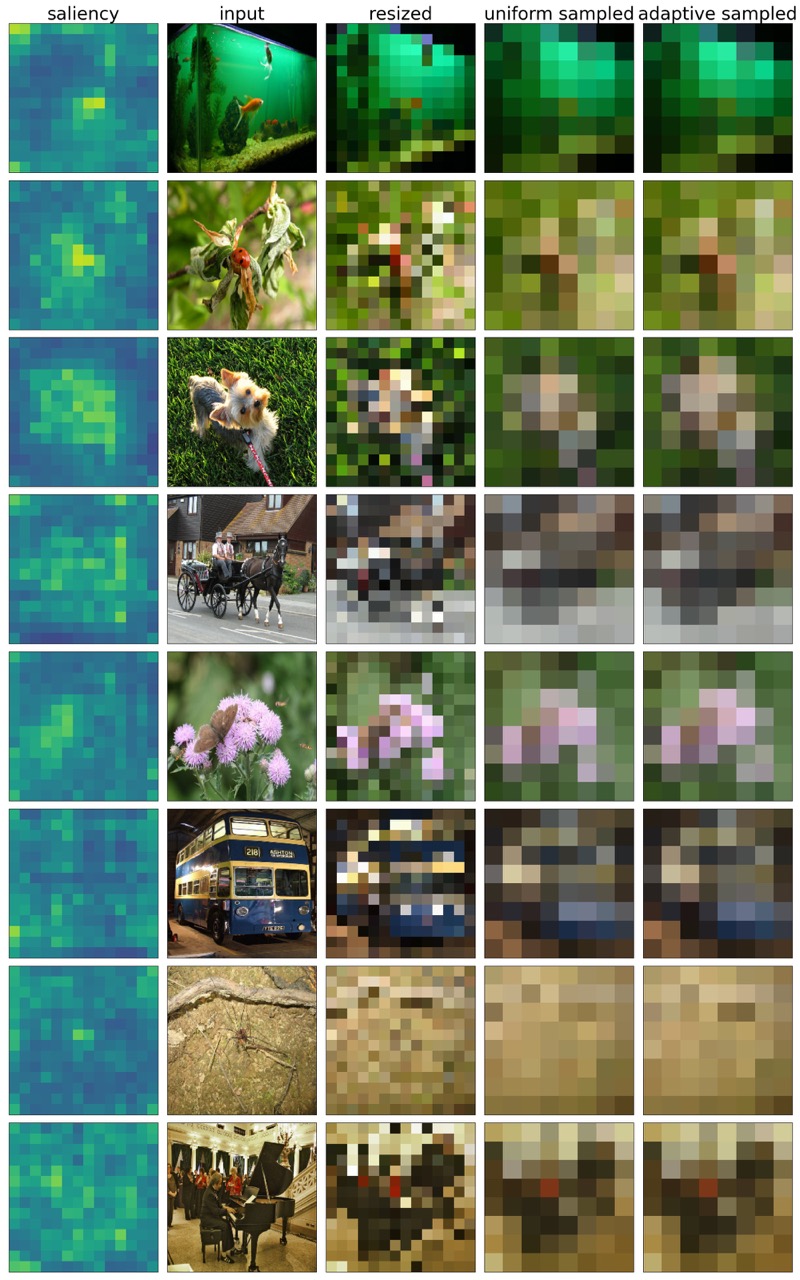}
		\caption{Example 1}
		\label{fig:visx2_4a}		
	\end{subfigure}
	\hfill
	\begin{subfigure}[t]{0.475\textwidth}
		\includegraphics[trim={0 5.6cm 0 0},clip,width=1.0\linewidth]{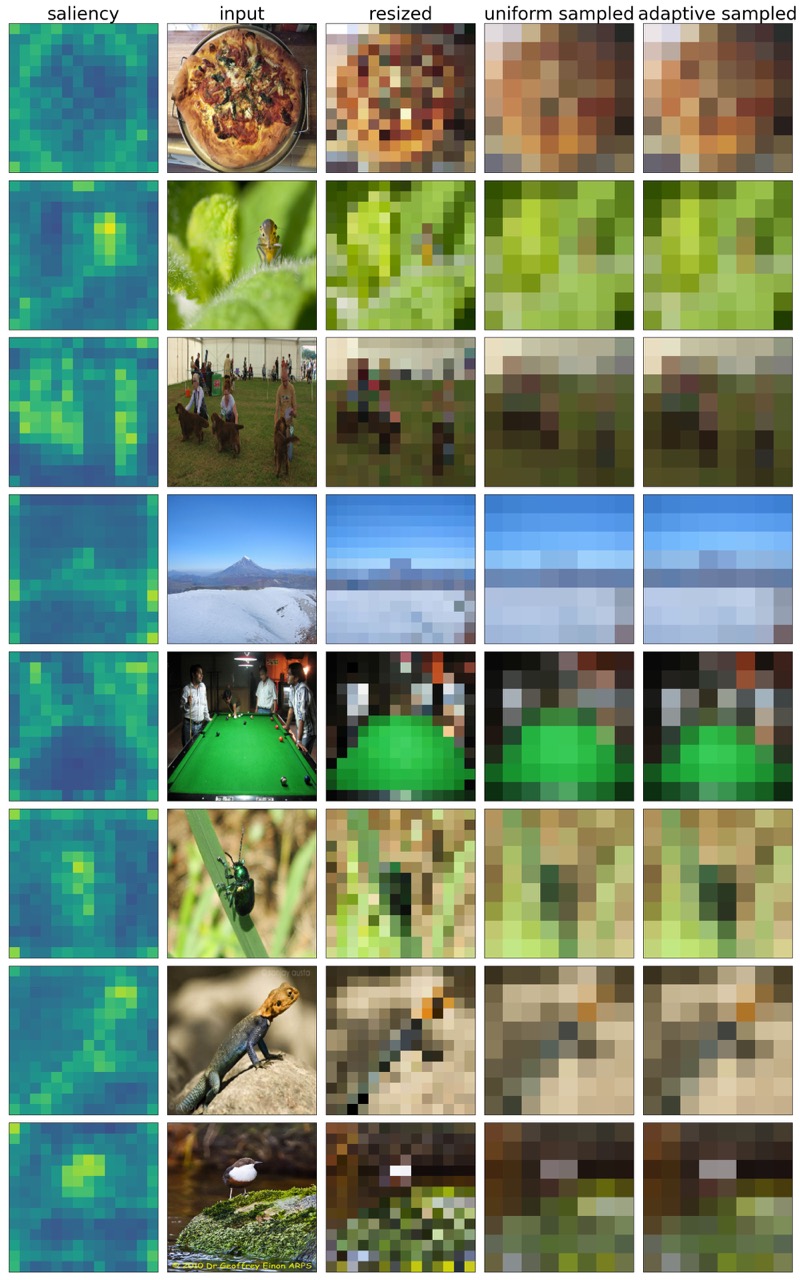}
		\caption{Example 2}
		\label{fig:visx2_4b}
	\end{subfigure}
	\caption{Layer 3-10}
	\label{fig:visx2_4}
\end{figure}

\begin{figure}[t]
	\begin{subfigure}[t]{0.475\textwidth}
		\includegraphics[trim={0 5.6cm 0 0},clip,width=1.0\linewidth]{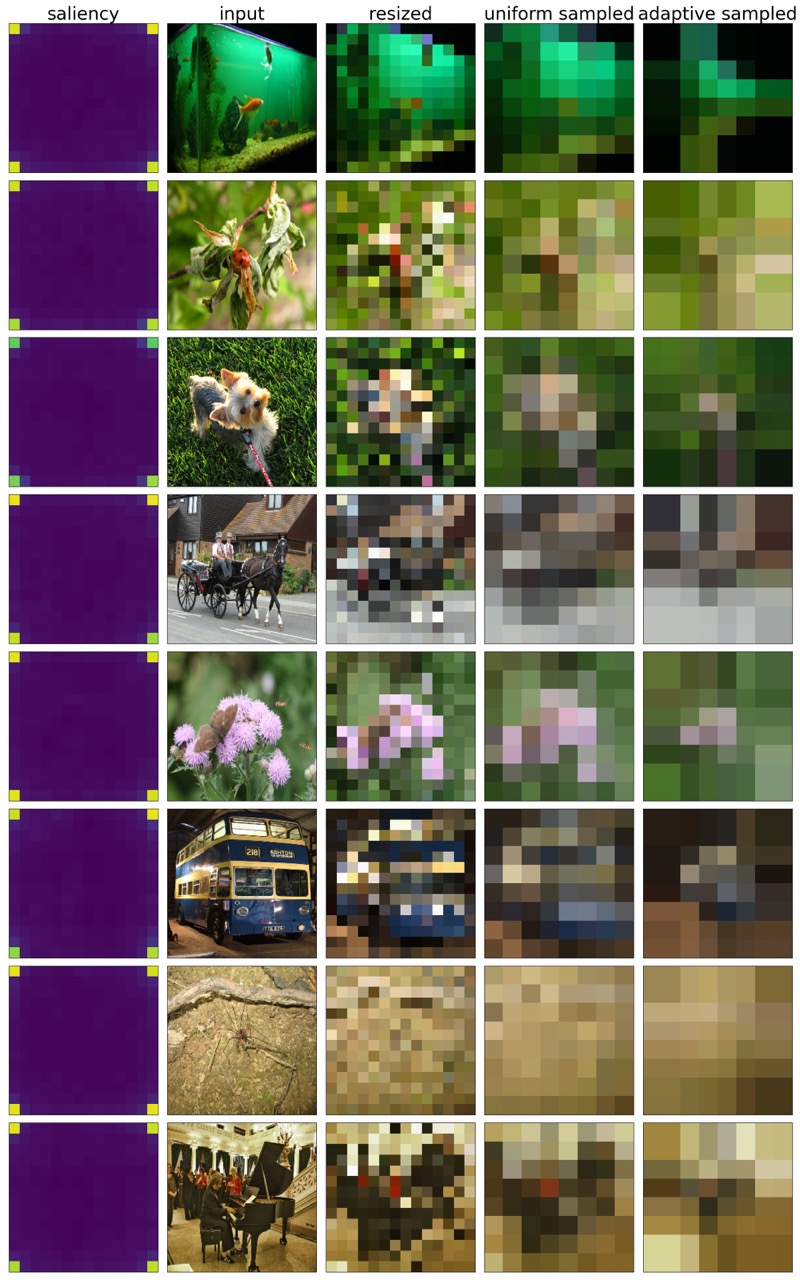}
		\caption{Example 1}
		\label{fig:visx2_5a}
	\end{subfigure}
	\hfill
	\begin{subfigure}[t]{0.475\textwidth}
		\includegraphics[trim={0 5.6cm 0 0},clip,width=1.0\linewidth]{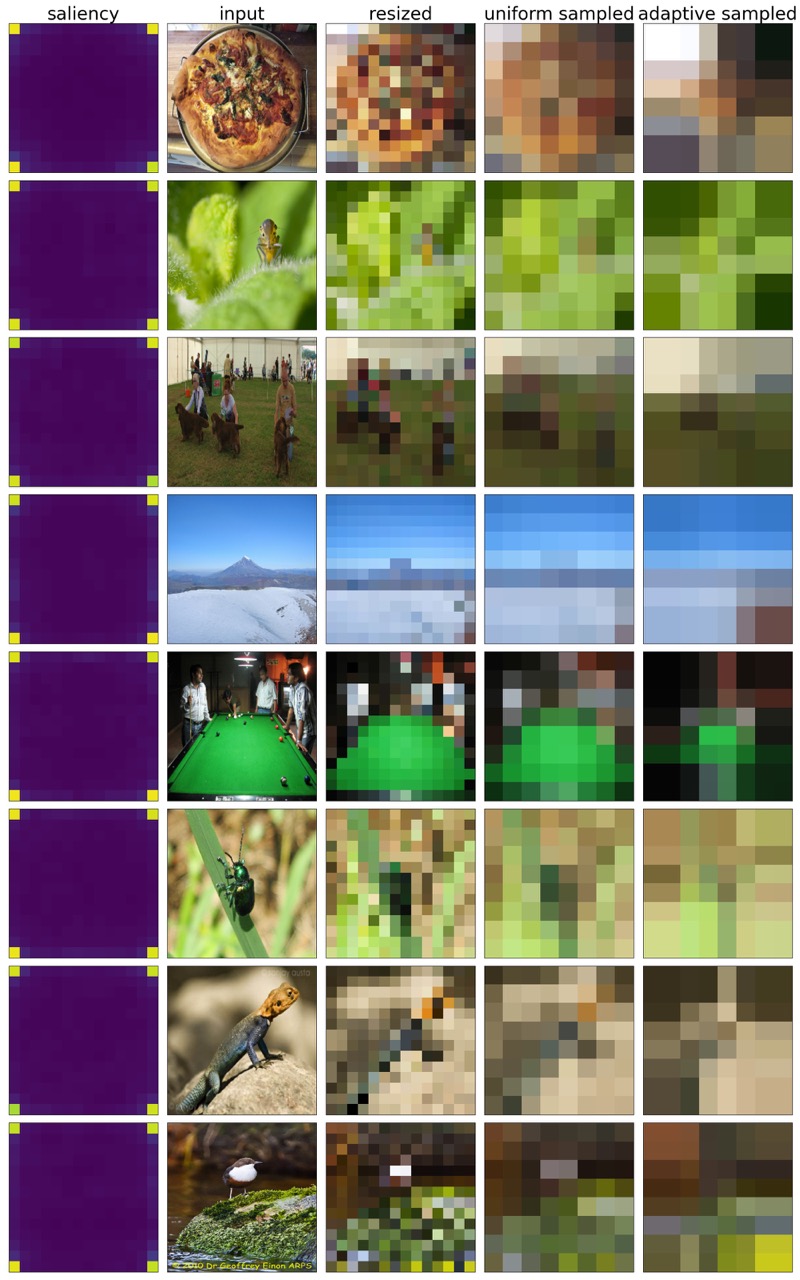}
		\caption{Example 2}
		\label{fig:visx2_5b}
	\end{subfigure}
	\caption{Layer 3-15}
	\label{fig:visx2_5}
\end{figure}

\begin{figure}[t]
	\begin{subfigure}[t]{0.475\textwidth}
		\includegraphics[trim={0 5.6cm 0 0},clip,width=1.0\linewidth]{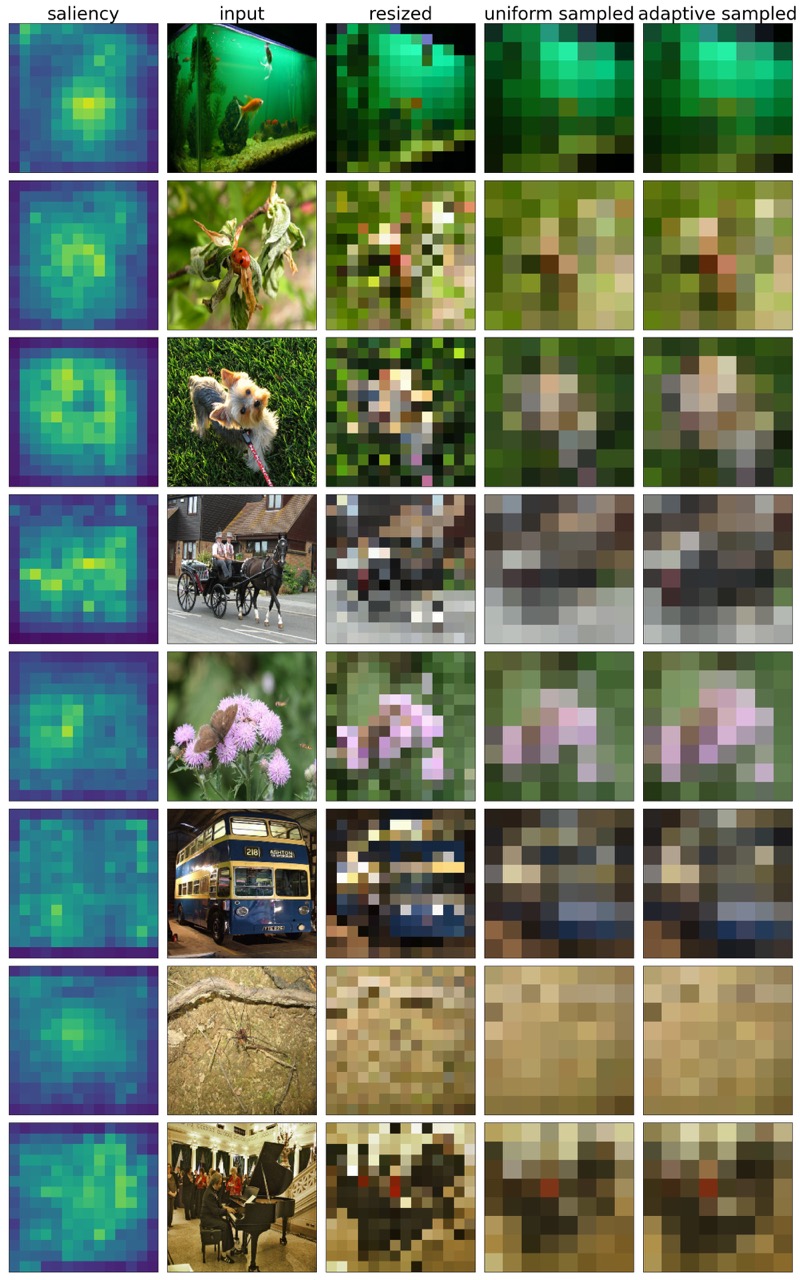}
		\caption{Example 1}
		\label{fig:visx2_6a}
	\end{subfigure}
	\hfill
	\begin{subfigure}[t]{0.475\textwidth}
		\includegraphics[trim={0 5.6cm 0 0},clip,width=1.0\linewidth]{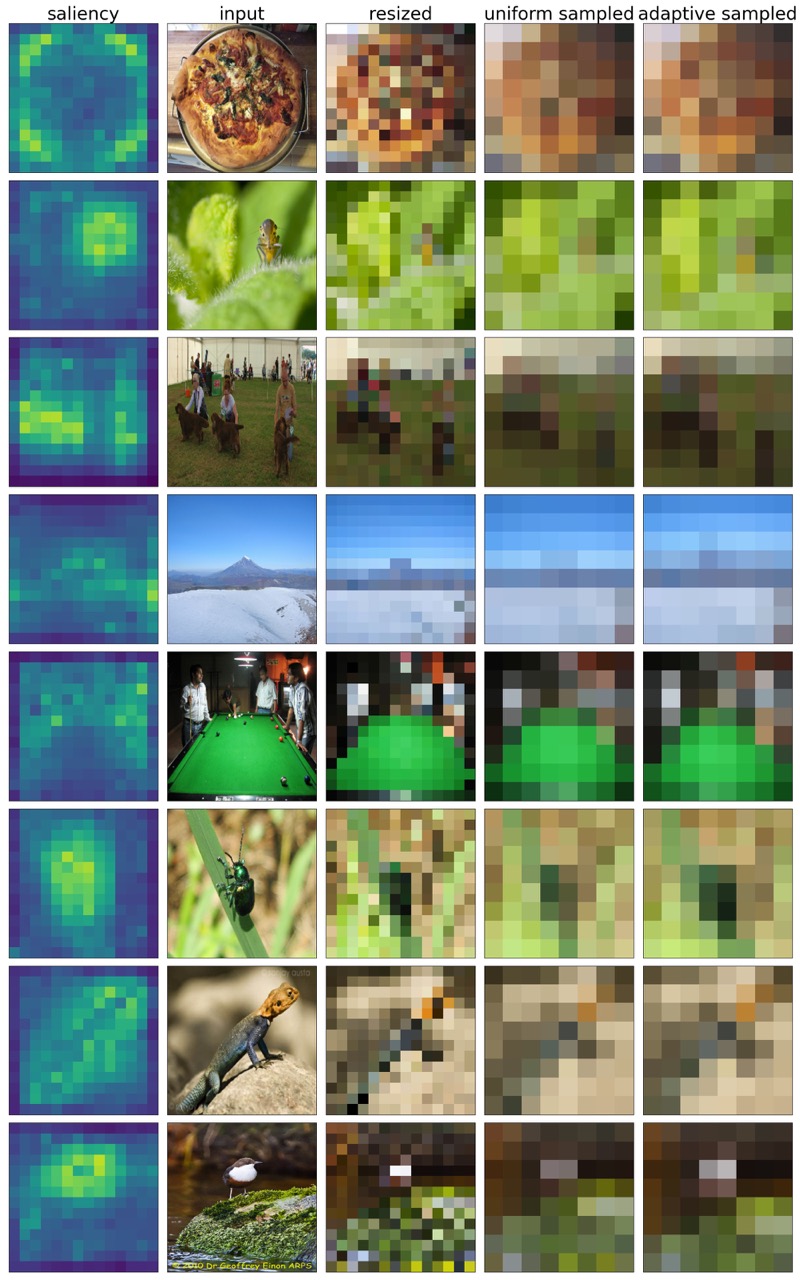}
		\caption{Example 2}
		\label{fig:visx2_6b}
	\end{subfigure}
	\caption{Layer 3-20}
	\label{fig:visx2_6}
\end{figure}

\begin{figure}[t]
	\begin{subfigure}[t]{0.475\textwidth}
		\includegraphics[trim={0 5.6cm 0 0},clip,width=1.0\linewidth]{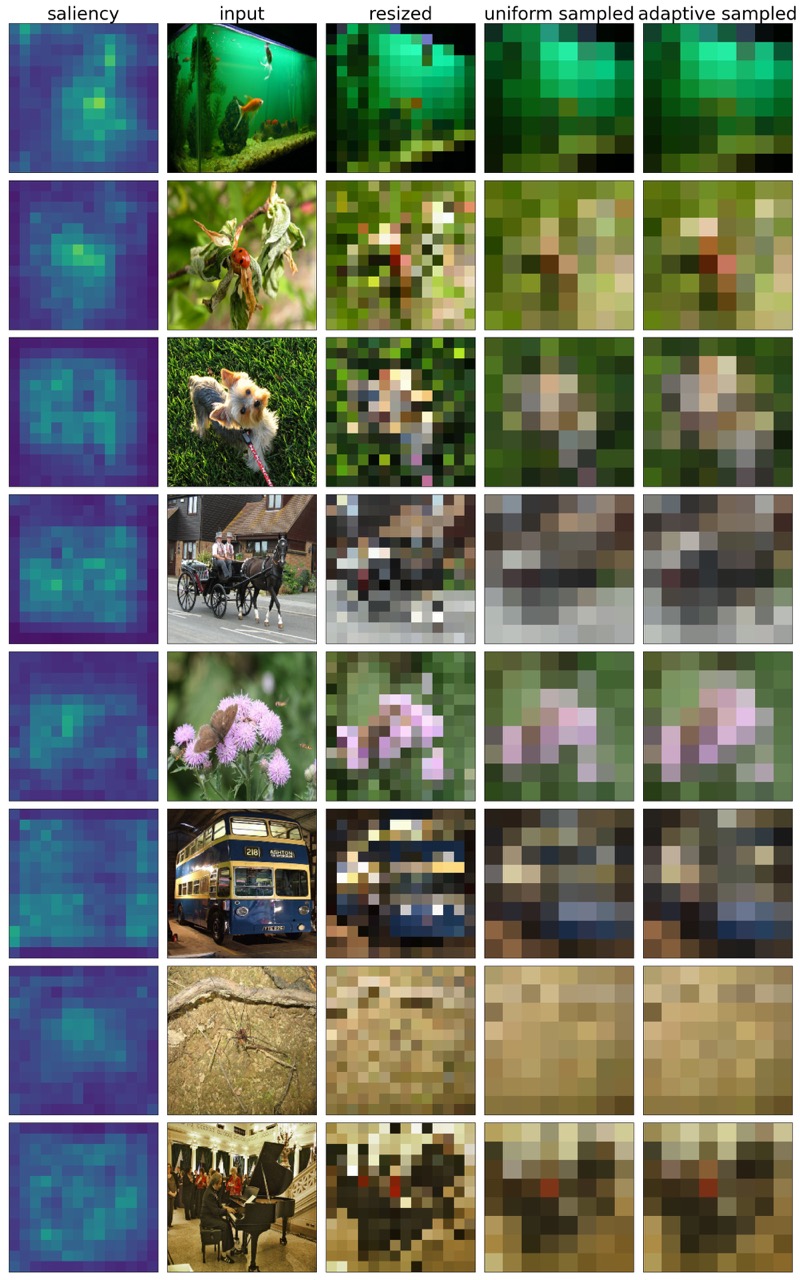}
		\caption{Example 1}
		\label{fig:visx2_7a}
	\end{subfigure}
	\hfill
	\begin{subfigure}[t]{0.475\textwidth}
		\includegraphics[trim={0 5.6cm 0 0},clip,width=1.0\linewidth]{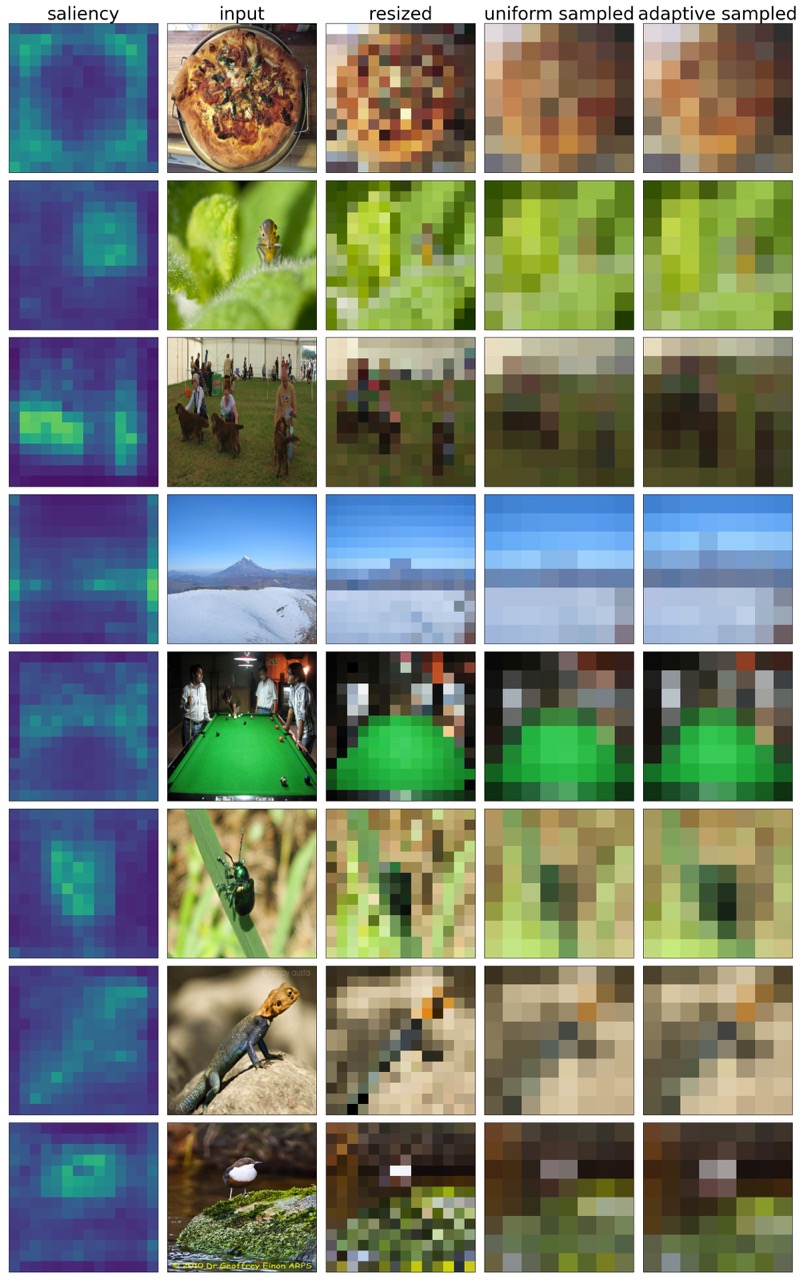}
		\caption{Example 2}
		\label{fig:visx2_7b}
	\end{subfigure}
	\caption{Layer 3-25}
	\label{fig:visx2_7}
\end{figure}

\begin{figure}[t]
	\begin{subfigure}[t]{0.475\textwidth}
		\includegraphics[trim={0 5.6cm 0 0},clip,width=1.0\linewidth]{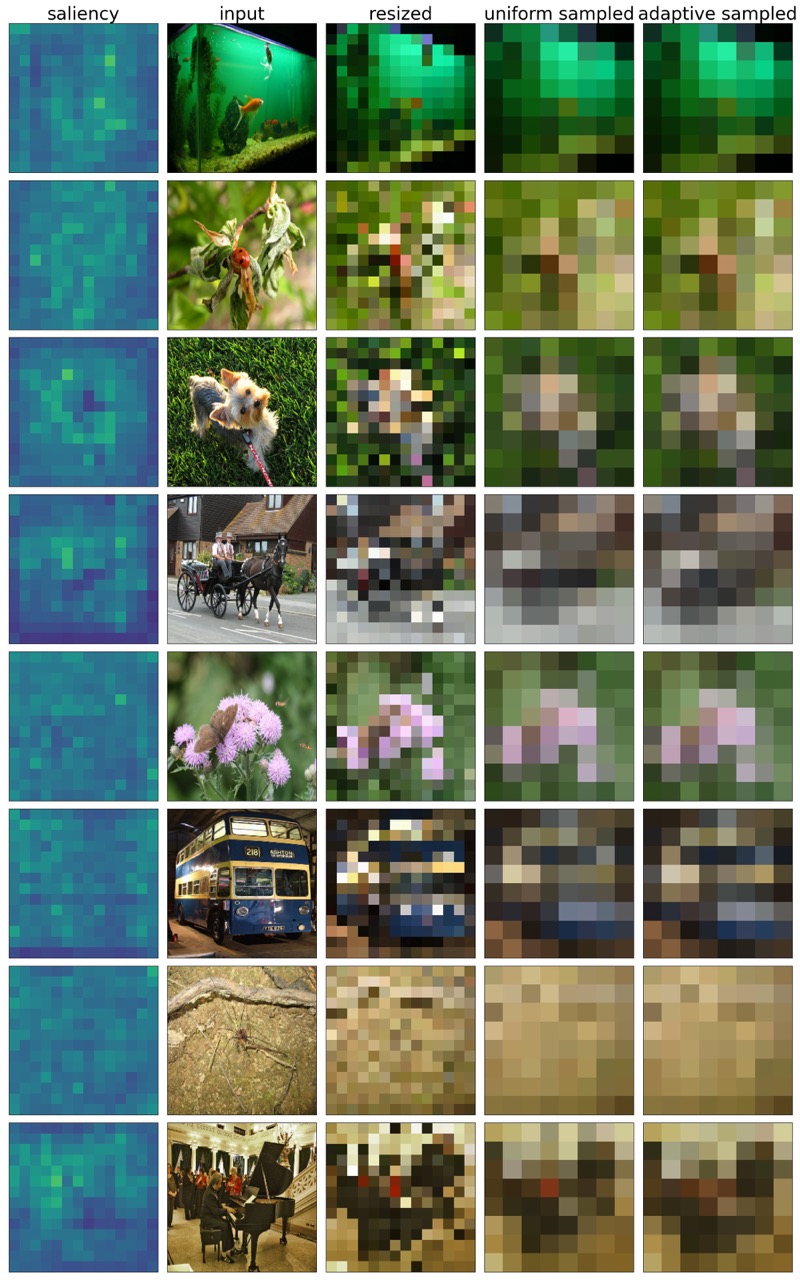}
		\caption{Example 1}
	\end{subfigure}
	\hfill
	\begin{subfigure}[t]{0.475\textwidth}
		\includegraphics[trim={0 5.6cm 0 0},clip,width=1.0\linewidth]{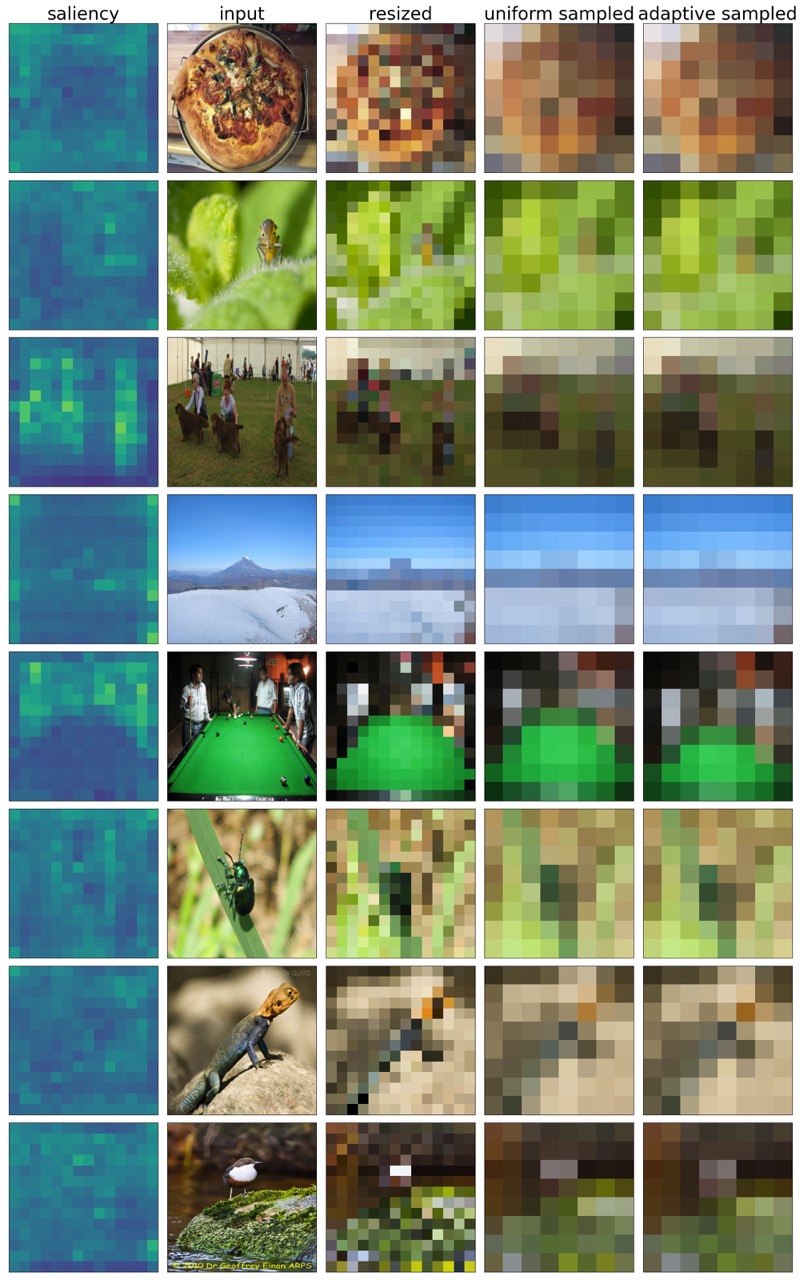}
		\caption{Example 2}
	\end{subfigure}
	\caption{Layer 3-30}
	\label{fig:visx2_8}
\end{figure}

\begin{figure}[t]
	\begin{subfigure}[t]{0.475\textwidth}
		\includegraphics[trim={0 5.6cm 0 0},clip,width=1.0\linewidth]{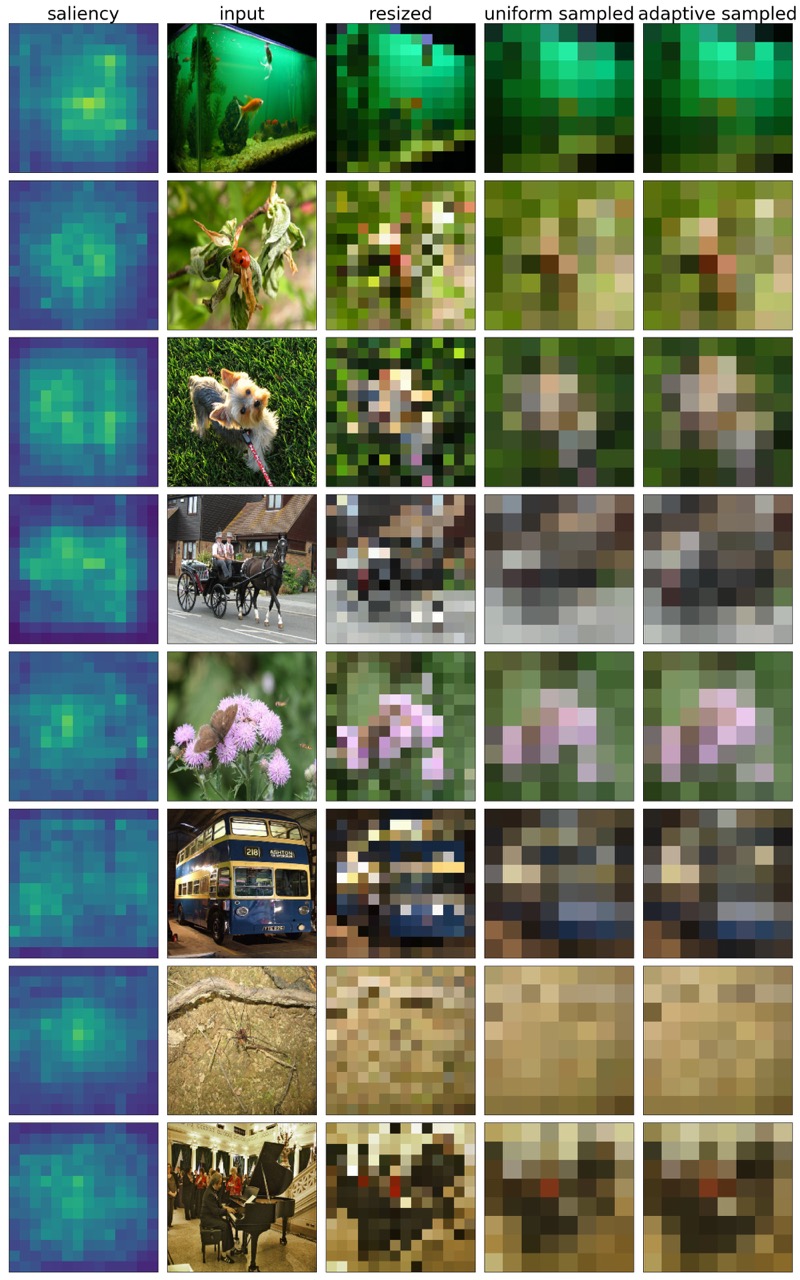}
		\caption{Example 1}
	\end{subfigure}
	\hfill
	\begin{subfigure}[t]{0.475\textwidth}
		\includegraphics[trim={0 5.6cm 0 0},clip,width=1.0\linewidth]{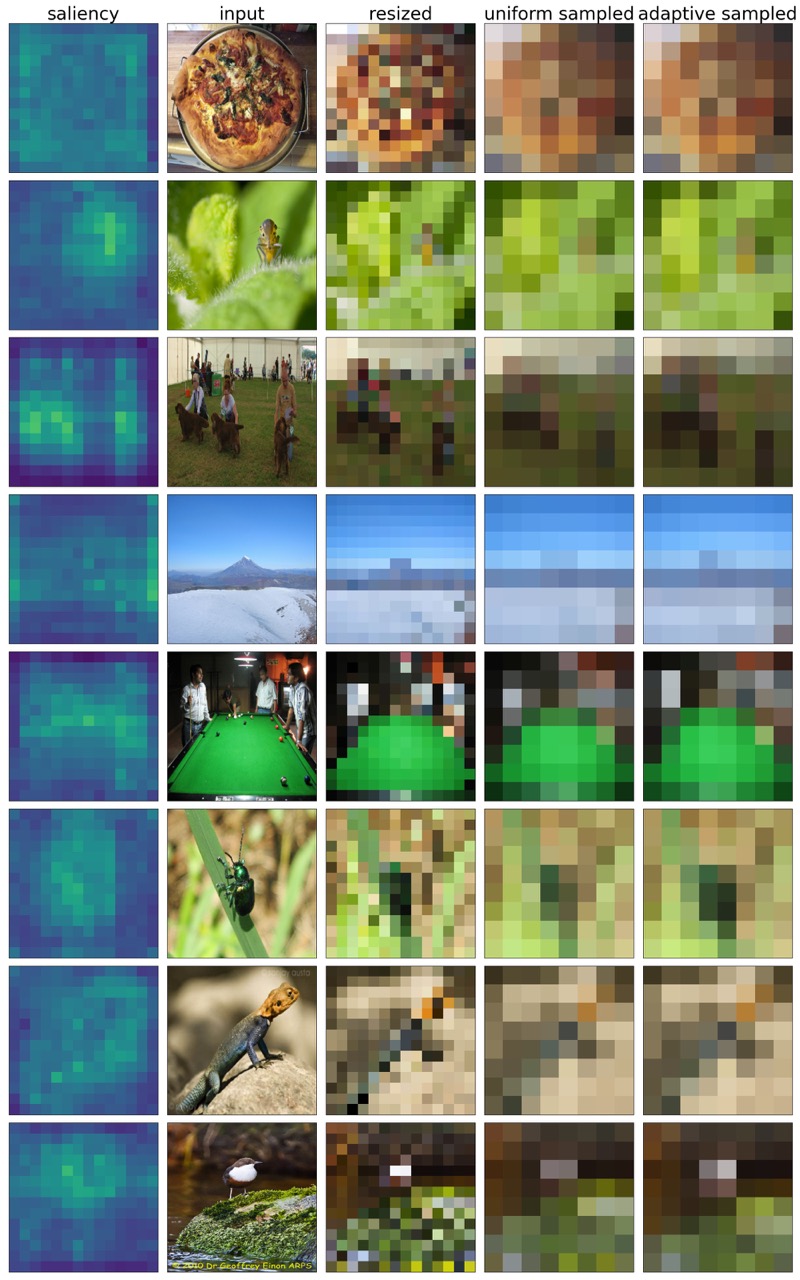}
		\caption{Example 2}
	\end{subfigure}
	\caption{Layer 3-35}
	\label{fig:visx2_9}
\end{figure}

\begin{figure}[t]
	\begin{subfigure}[t]{0.475\textwidth}
		\includegraphics[trim={0 5.6cm 0 0},clip,width=1.0\linewidth]{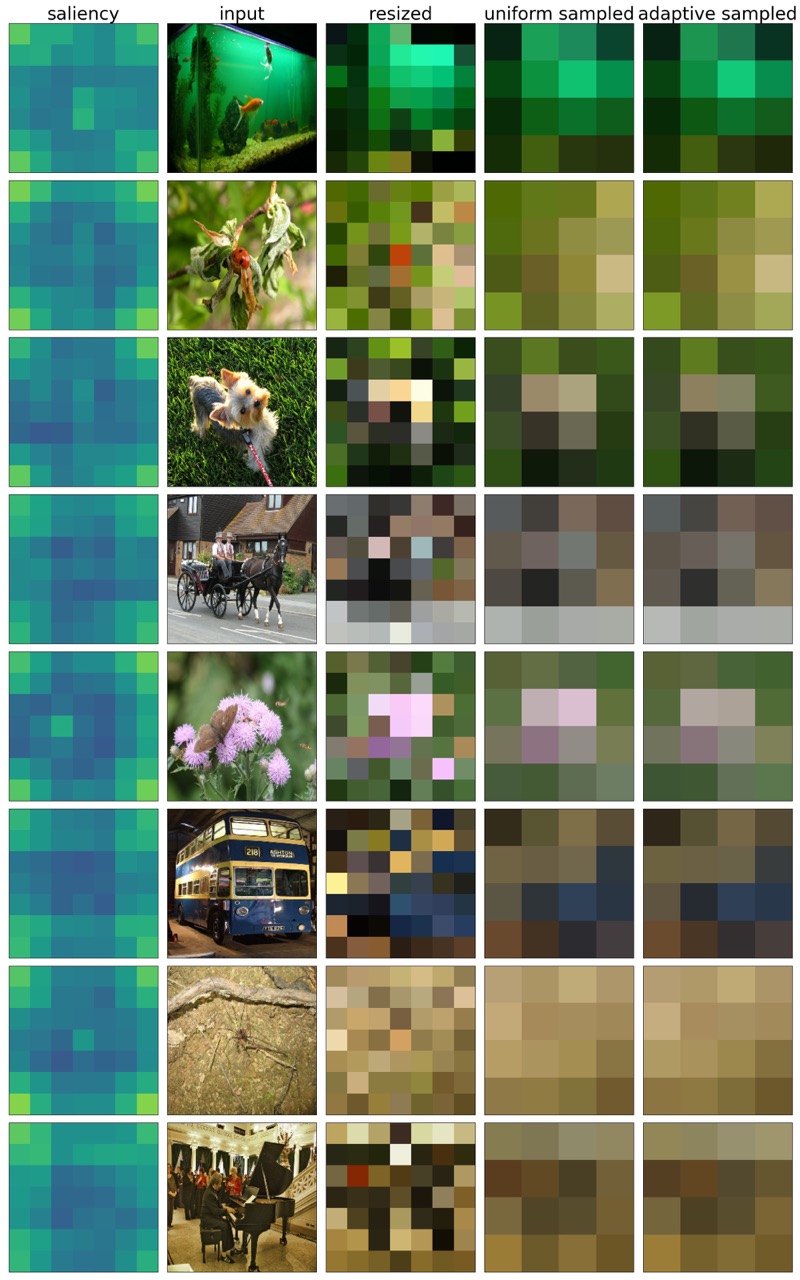}
		\caption{Example 1}
	\end{subfigure}
	\hfill
	\begin{subfigure}[t]{0.475\textwidth}
		\includegraphics[trim={0 5.6cm 0 0},clip,width=1.0\linewidth]{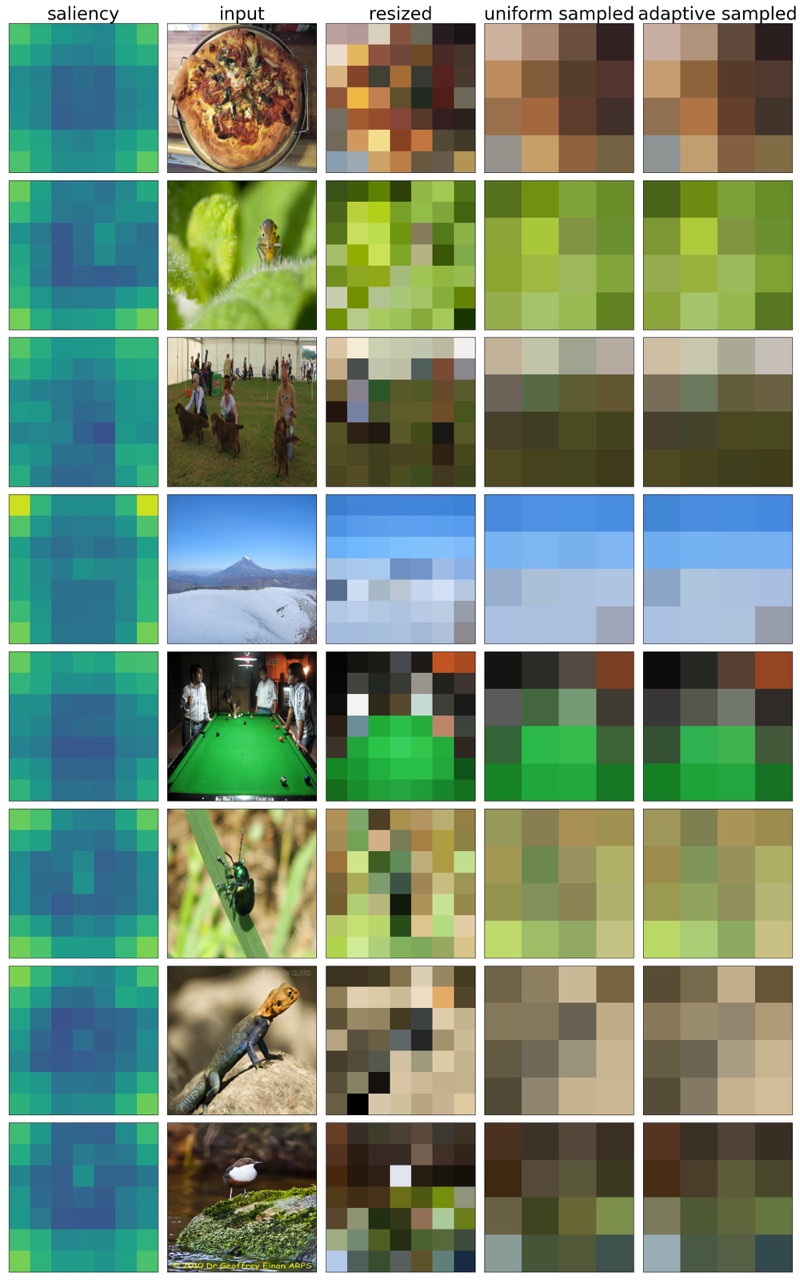}
		\caption{Example 2}
	\end{subfigure}
	\caption{Layer 4-3}
	\label{fig:visx2_10}
\end{figure}

To further illustrate the capability of SSBNet to focus on different locations of the feature maps at different layers, we provide additional visualization results in Figures \ref{fig:visx2_1}-\ref{fig:visx2_10}. All figures are sampled from the outputs of SSB-ResNet-RS-152 with input size of $224 \times 224$. The input images are from ImageNet dataset\cite{ILSVRC15}. Each figure provides the samples from one of the selected layers, and each row in the figures shows the results with different inputs. The figures are annotated with their index in the network, e.g. Layer 3-5 indicates the fifth layer in the third group of the building layers.

It can be observed that SSBNet is able to sample different positions at different layers. For example, in Figure \ref{fig:visx2_1}, the network samples the feature maps uniformly; in Figure \ref{fig:visx2_2}, it samples more heavier toward the objects; in Figure  \ref{fig:visx2_5}, the network zooms out from the feature maps, which increases the receptive field of the convolutional layers with respect to the input feature maps. We also noticed that the network weights different objects. In Figure \ref{fig:visx2_2b}, it focuses on both the people and the dogs in the third feature maps, while in Figure \ref{fig:visx2_7b}, it weights the dogs much heavier than the people.

\clearpage

\begin{table}[t]
	\scriptsize
	\setlength{\tabcolsep}{2pt}
	\caption{Comparison between uniform and adaptive sampling on ImageNet, with different sampling size}
	\begin{tabular}[t]{@{}lc c c c c}
		\toprule
		Model & Input size& Params & FLOPS & Top-1(\%) \\
		\midrule[1pt]
		U-50 & $224\times224$ & 25.6M & 2.61G & $77.70$ \\
		A-50 & $224\times224$ & 25.6M & 2.61G & $77.75$ \\
		\midrule
		U-101 & $224\times224$ & 44.6M & 3.35G & $78.29$ \\
		A-101 & $224\times224$ & 44.6M & 3.35G & $78.46$\\
		\midrule
		U-152 & $224\times224$ & 60.2M & 4.13G & $78.56$ \\
		A-152 & $224\times224$ & 60.3M & 4.14G & $78.64$ \\
		\midrule
		\multicolumn{5}{@{} l}{U: Uniform (12, 6, 3)} \\
		\multicolumn{5}{@{} l}{A: Adaptive (12, 6, 3)} \\
		\bottomrule
	\end{tabular}
	\hfill
	\begin{tabular}[t]{@{}lc c c c c}
		\toprule
		Model & Input size& Params & FLOPS & Top-1(\%) \\
		\midrule[1pt]
		U-50 & $224\times224$ & 25.6M & 2.95G & $77.96$  \\
		A-50 & $224\times224$ & 25.6M & 2.95G &  $78.09$ \\
		\midrule
		U-101 & $224\times224$ & 44.6M & 4.25G & $78.68$ \\
		A-101 & $224\times224$ & 44.6M & 4.25G & $78.88$\\
		\midrule
		U-152 & $224\times224$ & 60.2M & 5.60G & $79.05$ \\
		A-152 & $224\times224$ & 60.3M & 5.61G & $79.19$ \\
		\midrule
		\multicolumn{5}{@{} l}{U: Uniform (16, 8, 4)} \\
		\multicolumn{5}{@{} l}{A: Adaptive (16, 8, 4)} \\
		\bottomrule
	\end{tabular}
	\begin{tabular}[t]{@{}lc c c c c}
		\toprule
		Model & Input size& Params & FLOPS & Top-1(\%) \\
		\midrule[1pt]
		U-50 & $224\times224$ & 25.6M & 3.38G & $78.01$  \\
		A-50 & $224\times224$ & 25.6M & 3.38G &  $78.12$ \\
		\midrule
		U-101 & $224\times224$ & 44.6M & 5.40G & $79.01$ \\
		A-101 & $224\times224$ & 44.6M & 5.40G & $78.99$\\
		\midrule
		U-152 & $224\times224$ & 60.2M & 7.49G & $79.42$ \\
		A-152 & $224\times224$ & 60.3M & 7.49G & $79.45$ \\
		\midrule
		\multicolumn{5}{@{} l}{U: Uniform (20, 10, 5)} \\
		\multicolumn{5}{@{} l}{A: Adaptive (20, 10, 5)} \\
		\bottomrule
	\end{tabular}
	\hfill
	\begin{tabular}[t]{@{}lc c c c c}
		\toprule
		Model & Input size& Params & FLOPS & Top-1(\%) \\
		\midrule[1pt]
		B-50 & $224\times224$ & 25.6M & 2.95G & $77.62$  \\
		D-50 & $224\times224$ & 25.9M & 2.74G &  $77.95$ \\
		\midrule
		B-101 & $224\times224$ & 44.6M & 4.25G & $78.41$ \\
		D-101 & $224\times224$ & 45.3M & 3.69G & $78.42$\\
		\midrule
		B-152 & $224\times224$ & 60.2M & 5.61G & $78.86$ \\
		D-152 & $224\times224$ & 61.4M & 4.65G & $78.81$ \\
		\midrule
		\multicolumn{5}{@{} l}{B: Blinear (16, 8, 4)} \\
		\multicolumn{5}{@{} l}{D: DConv + Bilinear (14, 7, 4)} \\
		\bottomrule
	\end{tabular}
	\\
	\label{tab:imagenet_sampling}
\end{table}

\begin{table}[t]
	\scriptsize
	\setlength{\tabcolsep}{2pt}
	\caption{Comparison between uniform and adaptive sampling on ImageNet, with RandAugment\cite{cubuk2020randaugment}}
	\begin{tabular}[t]{@{}lc c c c c}
		\toprule
		Model & Input size& Params & FLOPS & Top-1(\%) \\
		\midrule[1pt]
		U-50 & $224\times224$ & 25.6M & 2.61G & $77.97$  \\
		A-50 & $224\times224$ & 25.6M & 2.61G &  $77.96$ \\
		\midrule
		U-101 & $224\times224$ & 44.6M & 3.35G & $78.91$ \\
		A-101 & $224\times224$ & 44.6M & 3.35G & $79.13$\\
		\midrule
		U-152 & $224\times224$ & 60.2M & 4.13G & $79.31$ \\
		A-152 & $224\times224$ & 60.3M & 4.14G & $79.65$ \\
		\midrule
		\multicolumn{5}{@{} l}{U: Uniform (12, 6, 3)} \\
		\multicolumn{5}{@{} l}{A: Adaptive (12, 6, 3)} \\
		\bottomrule
	\end{tabular}
	\hfill
	\begin{tabular}[t]{@{}lc c c c c}
		\toprule
		Model & Input size& Params & FLOPS & Top-1(\%) \\
		\midrule[1pt]
		U-50 & $224\times224$ & 25.6M & 2.95G & $78.07$ \\
		A-50 & $224\times224$ & 25.6M & 2.95G &  $78.22$ \\
		\midrule
		U-101 & $224\times224$ & 44.6M & 4.25G & $79.27$ \\
		A-101 & $224\times224$ & 44.6M & 4.25G & $79.50$\\
		\midrule
		U-152 & $224\times224$ & 60.2M & 5.60G & $79.71$ \\
		A-152 & $224\times224$ & 60.3M & 5.61G & $80.12$ \\
		\midrule
		\multicolumn{5}{@{} l}{U: Uniform (16, 8, 5)} \\
		\multicolumn{5}{@{} l}{A: Adaptive (16, 8, 4)} \\
		\bottomrule
	\end{tabular}
	\begin{tabular}[t]{@{}lc c c c c}
		\toprule
		Model & Input size& Params & FLOPS & Top-1(\%) \\
		\midrule[1pt]
		U-50 & $224\times224$ & 25.6M & 3.38G & $78.37$ \\
		A-50 & $224\times224$ & 25.6M & 3.38G &  $78.36$ \\
		\midrule
		U-101 & $224\times224$ & 44.6M & 5.40G & $79.41$ \\
		A-101 & $224\times224$ & 44.6M & 5.40G & $79.88$\\
		\midrule
		U-152 & $224\times224$ & 60.2M & 7.49G & $80.19$ \\
		A-152 & $224\times224$ & 60.3M & 7.49G & $80.28$ \\
		\midrule
		\multicolumn{5}{@{} l}{U: Uniform (20, 10, 5)} \\
		\multicolumn{5}{@{} l}{A: Adaptive (20, 10, 5)} \\
		\bottomrule
	\end{tabular}
	\hfill
	\begin{tabular}[t]{@{}lc c c c c}
		\toprule
		Model & Input size& Params & FLOPS & Top-1(\%) \\
		\midrule[1pt]
		B-50 & $224\times224$ & 25.6M & 2.95G & $77.77$ \\
		D-50 & $224\times224$ & 25.9M & 2.74G &  $78.06$ \\
		\midrule
		B-101 & $224\times224$ & 44.6M & 4.25G & $79.02$ \\
		D-101 & $224\times224$ & 45.3M & 3.69G & $79.15$ \\
		\midrule
		B-152 & $224\times224$ & 60.2M & 5.61G & $79.58$ \\
		D-152 & $224\times224$ & 61.4M & 4.65G & $79.58$ \\
		\midrule
		\multicolumn{5}{@{} l}{B: Blinear (16, 8, 4)} \\
		\multicolumn{5}{@{} l}{D: DConv + Bilinear (14, 7, 4)} \\
		\bottomrule
	\end{tabular}
	\\
	\label{tab:imagenet_sampling_ra}
\end{table}

\clearpage

\clearpage

\end{document}